\documentclass[letterpaper,twocolumn,10pt]{article}
\usepackage{usenix-2020-09}

\usepackage{tikz}
\usepackage{amsmath}
\usepackage{booktabs}
\usepackage{amssymb}

\usepackage{filecontents}

\begin{document}

\date{}

\title{\Large \bf Smaller is Better: Enhancing Transparency in Vehicle AI Systems via Pruning}

\author{
{\rm Sanish Suwal}\\
Rochester Institute of Technology
\and
{\rm Shaurya Garg}\\
Independent Researcher
\and
{\rm Dipkamal Bhusal}\\
Rochester Institute of Technology 
\and
{\rm Michael Clifford}\\
 Toyota InfoTech Labs
 \and
{\rm Nidhi Rastogi}\\
Rochester Institute of Technology
}

\maketitle

\begin{abstract}
Connected and autonomous vehicles continue to heavily rely on AI systems, where transparency and security are critical for trust and operational safety. Post-hoc explanations provide transparency to these black-box like AI models but the quality and reliability of these explanations is often questioned due to inconsistencies and lack of faithfulness in representing model decisions. This paper systematically examines the impact of three widely used training approaches, namely natural training, adversarial training, and pruning, affect the quality of post-hoc explanations for traffic sign classifiers. Through extensive empirical evaluation, we demonstrate that pruning \textit{significantly} enhances the comprehensibility and faithfulness of explanations (using saliency maps). Our findings reveal that pruning not only improves model efficiency but also enforces sparsity in learned representation, leading to more interpretable and reliable decisions. Additionally, these insights suggest that pruning is a promising strategy for developing transparent deep learning models, especially in resource-constrained vehicular AI systems.
\end{abstract}

\section{Introduction}
The increasing reliance on artificial intelligence (AI) in vehicles, such as autonomous cars, drones, and other connected systems, has transformed transportation by enabling capabilities like real-time navigation, vehicle-to-everything (V2X) communication, and intelligent decision-making~\cite{kuutti2020survey}. Despite the fact that AI in vehicles promises enhanced autonomy and safety, the inherent opacity or ``black-box'' nature of these models challenges transparency and trust, creating potential security and compliance risks in safety-critical settings.~\cite{ribeiro2016should}.

\begin{figure}[h]
    \centering
    \includegraphics[width=0.9\linewidth]{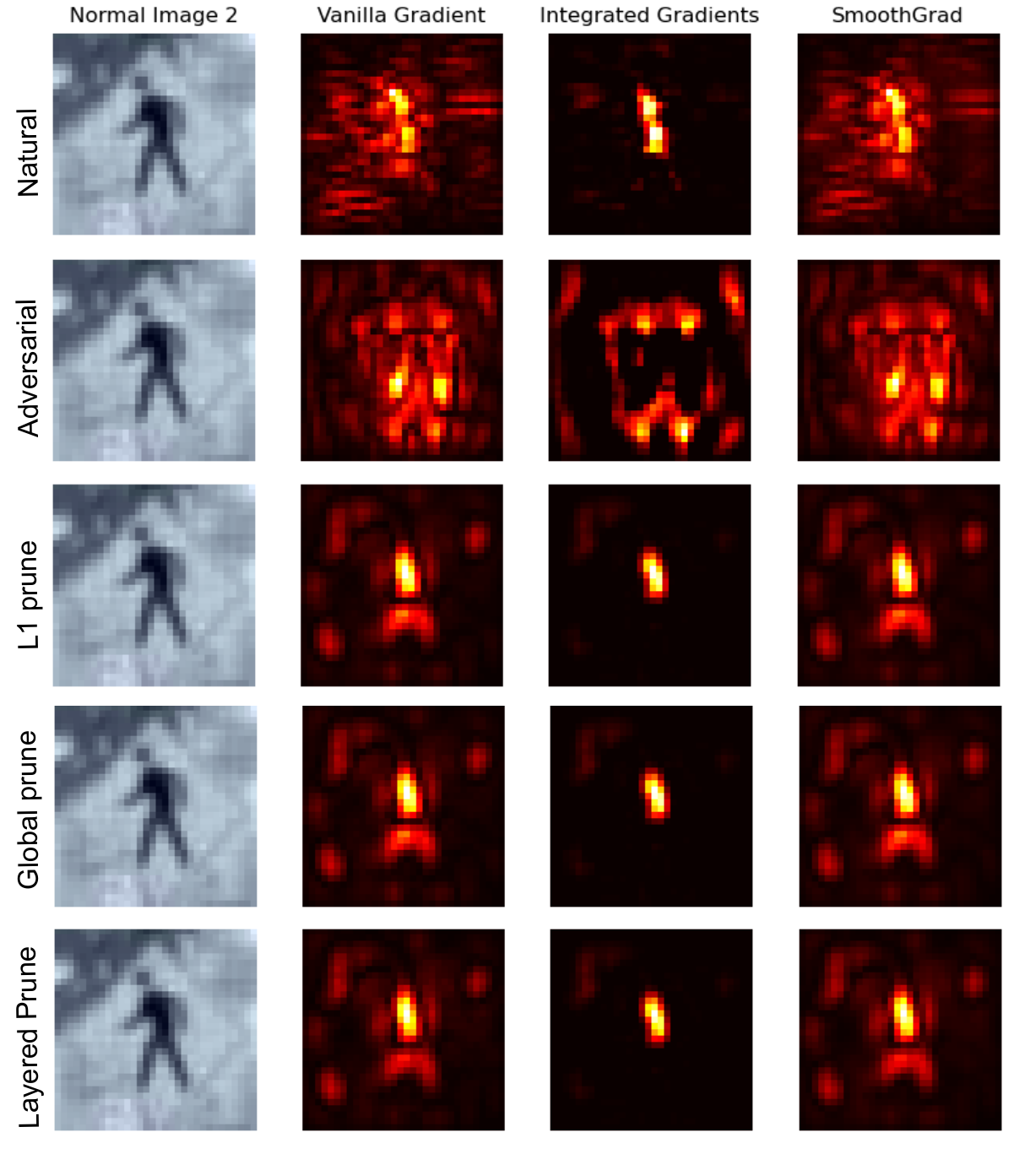}
    \caption{A figure showing heatmaps on an American traffic sign for naturally trained, adversarially trained and different pruned models for three explanation methods: Vanilla Gradient~\cite{simonyan2014deepinsideconvolutionalnetworks}, Integrated Gradient~\cite{sundararajan2017axiomaticattributiondeepnetworks} and SmoothGrad~\cite{smilkov2017smoothgrad}}
    \label{fig:intro-figure}
\end{figure}

In response to this challenge, research in explainable AI (XAI) has intensified, especially in developing \emph{post-hoc explanation} methods to foster trust, ensure compliance with safety standards, and facilitate debugging in adverse scenarios~\cite{bhusal2023sok}. Post-hoc explanation methods such as LIME~\cite{ribeiro2016should}, SHAP~\cite{lundberg2017unified}, Vanilla Gradient~\cite{simonyan2014deepinsideconvolutionalnetworks}, which interpret trained models by attributing importance to each input features and identifying the most influential ones. However, these methods are not without limitations. Perturbation-based methods like LIME and SHAP are shown to be unreliable in high-dimensional inputs like images~\cite{slack2020fooling}, and several explanation methods produce unfaithful explanations~\cite{adebayo2018sanity}. Similarly, gradient-based methods are sensitive to hyperparameter choices and frequently fail standard interpretability tests, leading to misleading explanations~\cite{adebayo2018sanity,bansal2020sam}.

Given these limitations, it is critical that post-hoc explanations are reliable, trustworthy, and comprehensible. In addition, they should ensure their practical utility by satisfying some key quantitative properties: sparsity~\cite{chalasani2020concise} and faithfulness~\cite{rong22consistent}. \textit{Sparsity} measures if the explanations focus on the most relevant features by discarding irrelevant ones, thus improving comprehensibility. \textit{Faithfulness} property requires that the post-hoc explanations accurately reflect the model’s actual decision-making process, thus explaining what the model actually did in making the prediction. While methods like Guided Backpropagation (GBP)\cite{springenberg2014striving}, Integrated Gradient (IG)~\cite{sundararajan2017axiomaticattributiondeepnetworks}, SmoothGrad~\cite{smilkov2017smoothgrad}, NoiseGrad~\cite{bykov2022noisegrad} have been proposed to improve the quality of explanations, they remain sensitive to change in their hyperparameters~\cite{bansal2020sam}, and fail several sanity tests~\cite{adebayo2018sanity, kindermans2019reliability}.

In this work, we take a complementary approach for improving the quality of post-hoc explanations. Recent works have shown that adversarial training~\cite{goodfellow2015explainingharnessingadversarialexamples} can improve the comprehensibility of saliency maps by encouraging models to rely on more robust and semantically meaningful features in high-dimensional images~\cite{etmann2019connection}. However, adversarial training is computationally expensive, requiring generation of adversarial samples and additional training overhead often leading to performance degradation on benign inputs. This raises a question: \textbf{Can we achieve similar improvements in explanation quality without the drawbacks of adversarial training?}


Pruning~\cite{han2015learning}, traditionally used for model compression, offers an alternative. By removing less significant parameters, pruning not only reduces model complexity but also enforces sparsity in learned representations. We hypothesize that this sparsity can significantly improve the interpretability and reliability of saliency-based explanations by highlighting critical decision-making features. Unlike adversarial training, pruning is lightweight, can be applied both pre- \textit{and} post-training, and is more suitable for deployment in resource-constrained environments like autonomous vehicles.

\textit{To address this gap}, we systematically analyze the impact of three widely used training approaches: \textit{natural, adversarial and pruning}. We measure saliency map quality for traffic sign recognition models trained on multiple datasets, namely the American traffic sign dataset, LISA~\cite{GTDLBenchICDCS} and the German traffic sign dataset, GTSRB~\cite{stallkamp2012man}. Our extensive empirical evaluation reveals that pruning enhances the comprehensibility of saliency maps, which improves their faithfulness, thereby reflecting the underlying decision-making of the model. Figure \ref{fig:intro-figure} shows saliency maps on a traffic sign for different models using three explanation methods, where we can clearly observe that pruning improves the interpretability of saliency maps by focusing on critical features of the input image.

\textbf{Key contributions:} While pruning is primarily used for model compression and efficiency, its role in enhancing interpretability remains unclear. This work aims to inspect how pruning impacts the sparsity and faithfulness of saliency maps by conducting extensive experiments on models trained on the LISA and GTSRB traffic sign datasets. 

Our main contributions are as follows:

\begin{enumerate}
    \item \textbf{Comprehensive evaluation of explanations:} In Section \ref{sec:methodology}, we systematically compare natural training, adversarial training, and different pruning techniques to assess their impact on saliency map interpretability of vehicular AI model. We perform both quantitative and qualitative evaluations that provide an in-depth assessment of explanation quality for different training strategies.

\item \textbf{New insights on pruning for interpretability:} Our results in Section \ref{sec:results} demonstrate that pruning enhances faithfulness and sparsity of saliency maps, making them more human-comprehensible and reliable. While adversarial training improves robustness of models, it often leads to noisier explanations, particularly in low-resolution datasets like LISA and GTSRB. This suggests that pruning, in addition to model compression, can be a more effective approach to improving interpretability without sacrificing model performance.

\item \textbf{Guidelines for model interpretability enhancement:} By demonstrating that pruning enhances interpretability, we provide an alternative pathway to model transparency. Pruning is a technique that is primarily used for model compression, however, its advantage in model transparency as demonstrated in our work, makes pruning particularly valuable in safety-critical scenarios, where human-comprehensible explanations are necessary for regulatory compliance and trustworthiness. 
\end{enumerate}

The rest of the paper is structured as follows: Section \ref{sec:background} describes the background on explainable AI,  explanation methods and their evaluation metrics, and different model training strategies. Section \ref{sec:methodology} discusses the methodology of the experiment that includes datasets, model training, and metrics. Section \ref{sec:results} presents the qualitative and quantitative evaluation of saliency maps, and discusses key findings, and implications. Section \ref{sec:conclusion} concludes the paper with recommendations for model selection and interpretability enhancements.

\section{Background and Related Work}\label{sec:background}
\subsection{Explainable AI- Methods and Challenges}\label{sec:xai}

Interpretability in deep learning (DL) models focuses on two primary research directions: (1) designing intrinsically interpretable models, and (2) developing post-hoc explanation methods. While the former focuses on building models whose structure and decisions are inherently explainable, the latter involves techniques that provide insights into the decisions of complex, opaque models by analyzing their predictions in relation to input features~\cite{bhusal2023sok}. Building inherently interpretable models with high performance for complex datasets like images, videos and texts is a challenging task, hence, post-hoc methods have gained significant traction, which have been demonstrated lately to help with detecting adversarial attacks~\cite{bhusal2024pasa, yang2020ml, wang2020interpretability}.  

\textbf{Post-hoc explanation methods} can be categorized based on several factors, including the granularity of explanations (local vs. global), the type of models they support (model-agnostic vs. model-specific), and the nature of the explanations they provide (feature attribution, rule-based, or counterfactual explanations). \textbf{Feature attribution methods} assign relevance scores to individual features, indicating their importance in the model's prediction for a given instance. Such scores are visualized as saliency maps in image classifiers. Formally, given a black-box model \( F(.) \) and a test instance \( \mathbf{x} = \{x_1, x_2, \dots, x_N\} \), where \( N \) represents the number of input features, a feature attribution method returns a vector \( \phi(\mathbf{x}) \) with the same dimension as \(\mathbf{x}\) that provides the relevance of each feature to the given model prediction \( F(\mathbf{x}) \). \textbf{Perturbation-based feature attribution methods} such as LIME~\cite{ribeiro2016should} perturb a given instance to generate multiple samples, train a simpler interpretable model (eg. logistic regression) to explain the test instance. \textbf{Backpropagation-based attribution methods} (e.g, Integrated Gradient~\cite{sundararajan2017axiomaticattributiondeepnetworks}) utilize gradients to propagate the model prediction to the input layer. These methods pass several sanity checks~\cite{adebayo2018sanity}, are faster to compute and widely used due to their versatility in applications and ease of interpretation~\cite{bhusal2023sok}. Hence, we focus on three representative gradient-based post-hoc explanation method: Vanilla Gradient (VG), Integrated Gradient (IG) and SmoothGrad (SG). 

\textbf{Vanilla Gradient (VG)~\cite{simonyan2014deepinsideconvolutionalnetworks}:} VG is the one of the earliest gradient-based explanation approach. This technique computes the gradient of the class score with respect to the input features. This gradient value indicates the sensitivity of the output prediction to changes in input features and is used as a feature attribution score, visualized as saliency maps (heatmap) for images.  

\begin{equation}\label{eqn:VGattribution}
    VG(\mathbf{x}) = \frac{\partial F(\mathbf{x})}{\partial {x_i}}
\end{equation}

Here, \(F(\mathbf{x})\) is the model's output function, and \(x_i\) represents an input feature. While simple and computationally efficient, Vanilla Gradient method can produce noisy explanations~\cite{smilkov2017smoothgrad}, especially for complex neural networks, and suffers from gradient saturation~\cite{shrikumar2017learning}.
\par
\textbf{Integrated Gradients (IG)~\cite{sundararajan2017axiomaticattributiondeepnetworks}:} IG builds upon Vanilla Gradient to address its limitations, particularly the issue of noisy attributions due to gradient saturation. IG introduces the concept of a baseline input (e.g., a black image for image-based models) and computes gradients along a straight path from this baseline to the actual input. By integrating these gradients, IG determines the contribution of each input feature to the output while satisfying two key axioms:  
\begin{enumerate}
    \item \textit{Sensitivity:} If a single input feature differs between the baseline and the input, and this difference affects the model's prediction, the attribution for that feature must be non-zero. 

    \item \textit{Implementation Invariance:} Two functionally equivalent models (even with different implementations) should yield identical attributions for the same input.
\end{enumerate}

These axioms ensure that IG provides robust and reliable attributions, addressing deficiencies in earlier methods. The IG formulation is given as:  

\begin{equation}\label{suppeqn:igequationsimple}
    IG_i ({x}) = (x_i - x'_i). \int_{\alpha=0}^{1} \frac{\partial F(\mathbf{x'}+\alpha (\mathbf{x} - \mathbf{x'}))}{\partial x_i} \partial \alpha 
\end{equation}

Here, \(\mathbf{x}'\) is the baseline, \(\mathbf{x}\) is the input, and \(\alpha\) controls the transition between them, varying from 0 to 1. This method has become one of the most widely used gradient-based approaches for feature attribution. The selection of an appropriate baseline depends on the context and dataset~\cite{sturmfels2020visualizing}, and can significantly influence the quality of the explanations.

\textbf{SmoothGrad (SG)~\cite{smilkov2017smoothgrad}:} To further address the noise in gradient-based saliency maps, Smilkov et al. introduced SmoothGrad. This method improves explanation stability by creating \(n\) perturbed versions of the input through the addition of Gaussian noise (\(N(0, \sigma^2)\)). Gradients are computed for each perturbed sample, and their average is taken to produce the final attribution map. This averaging reduces noise, highlights meaningful patterns, and generates smoother, more interpretable explanations. The formulation for SmoothGrad is as follows:  
 
\begin{equation}\label{eqn:SG}
    SG_i({x}) = \frac{1}{N} \sum_{k=1}^{N} \frac{\partial F(\mathbf{x} + N(0, \sigma^2))}{\partial {x}_i} 
\end{equation}

Here, \(N(0, \sigma^2)\) represents Gaussian noise. SmoothGrad often outperforms basic gradient-based methods by suppressing spurious artifacts in explanations.

\subsection{Evaluating Explanation Quality}\label{sec:metrics}

Qualitative evaluation of saliency maps are subjected to human-bias and makes it difficult to judge whether an explanation is correct. Quantitative evaluation of saliency maps, however, utilize formal definitions and properties of explanation quality, and does not require human validation. Below, we explain the evaluation metrics of saliency maps used in our evaluation:

\subsubsection{Measuring Explanation Sparsity} 

Sparsity measures how focused an attribution vector \( \phi(\textbf{x}) \) is by evaluating the distribution of importance across input features. We use the Gini index to quantify sparsity~\cite{chalasani2020concise}. For an attribution vector \( \phi(\textbf{x}) \in \mathbb{R}^d \), the elements are first sorted in non-decreasing order, and the Gini index is computed as follows:

\begin{equation}\label{eqn:gini}
    G(\phi(\mathbf{x})) = 1 - 2 \sum_{k=1}^d \frac{\phi(\mathbf{x})_{(k)}}{||\phi(\mathbf{x})||_1} \frac{d-k+0.5}{d}
\end{equation}

Here, \( ||\phi(\textbf{x})||_1 \) is the \( L_1 \)-norm of \( \phi(\textbf{x}) \), and \( \phi(\textbf{x})_{(k)} \) denotes the \( k \)-th smallest element in the sorted vector. The Gini index ranges from 0 to 1. A value of 1 indicates perfect sparsity, where only one element in \( \phi(\textbf{x}) \) is non-zero. A value of 0 indicates uniform distribution across all features. 

Sparsity helps evaluate how concentrated the attribution scores are, with higher sparsity leading to more comprehensible and human-friendly explanations.

\subsubsection{Measuring Explanation Faithfulness} 
Faithfulness measures the correctness of an explanation method in capturing relevant features for a given test sample. This is the most crucial quantitative metric as we want explanations to truly represent the model we want to explain. We measure faithfulness using ROAD metric~\cite{rong22consistent}. ROAD (Remove and Debias) evaluates the accuracy of a model as the most important features are iteratively removed. Using the MoRF (Most Relevant First) removal strategy, features are ordered by decreasing importance based on an attribution method, and the top \( k \) features are removed in each iteration. The accuracy is tracked at each step.

Given a model \( F \), input sample \( \mathbf{x} \), and an attribution method that assigns importance scores to features, the removal process is performed iteratively. The removal uses noisy linear imputation to prevent out-of-distribution samples. For our experiments, we set \( k=5 \).

A sharper accuracy drop as features are removed indicates a better explanation, as the most relevant features have a greater impact on model predictions. ROAD is preferred over other faithfulness metrics like Insertion/Deletion~\cite{petsiukrise} or ROAR~\cite{hooker2019benchmark}, as it avoids distribution shifts caused by perturbations and does not require expensive model retraining.

\subsection{Model training strategies}


\subsubsection{Natural Training}  

Natural training refers to the conventional optimization process where a model is trained by minimizing a loss function \( L \), typically using gradient descent or its variants~\cite{goodfellow2016deep}. For a model \( F \), input \( \mathbf{x} \), true label \( y \), and parameters \( \theta \), the objective is to find \( \theta \) that minimizes the empirical risk:  

\begin{equation}
\theta^* = \arg\min_\theta \mathbb{E}_{(x, y) \sim D} \big[ L(F(\mathbf{x}; \theta), y) \big]
\end{equation}

Here, \( D \) represents the data distribution. In this framework, the model learns to generalize by optimizing performance on the training data without any explicit mechanisms to handle robustness or interpretability. While effective in many scenarios, natural training often results in over-parameterized networks susceptible to adversarial attacks~\cite{goodfellow2015explainingharnessingadversarialexamples} and challenges in post-hoc explanation quality due to noise and saturation in gradients~\cite{smilkov2017smoothgrad}.

\subsubsection{Adversarial Training}

Adversarial training enhances a model’s robustness against adversarial attacks by training it on perturbed inputs~\cite{goodfellow2015explainingharnessingadversarialexamples}. Given an input \( \mathbf{x} \) and its true label \( y \), an adversarial example \( \mathbf{x}' \) is crafted to maximize the model’s loss:  

\begin{equation}
    \mathbf{x}' = \arg\max_{\mathbf{x}' \in B_\epsilon(\mathbf{x})} L(F(\mathbf{x}'; \theta), y)
\end{equation}

Here, \( B_\epsilon(\mathbf{x}) \) is an \( \epsilon \)-ball around \( x \), representing the set of valid perturbations constrained by \( ||\mathbf{x}' - \mathbf{x}|| \leq \epsilon \). Adversarial training involves minimizing the worst-case loss:  

\begin{equation}
\theta^* = \arg\min_\theta \mathbb{E}_{(\mathbf{x}, y) \sim D} \big[ \max_{\mathbf{x}' \in B_\epsilon(x)} L(F(\mathbf{x}'; \theta), y) \big]
\end{equation}

This approach improves the model’s resilience to malicious inputs, such as spoofed sensor data in vehicular systems. Recent studies have shown that post-hoc explanations of adversarially trained models are clearer and sparser than naturally trained models~\cite{etmann2019connection, zhang2019interpreting, chalasani2020concise}. However, its impact on explanation quality of vehicular system has not been thoroughly explored.

\subsubsection{Neural Network Pruning}\label{sec:pruning}

Large neural networks consume more memory space, require more time for computation, and present challenge for deployment on devices where computational resources is limited such as autonomous driving~\cite{han2015learning, han2015deep, dong2017more}. Pruning is an effective technique of compressing neural network to save memory space and inference time computation. It reduces the number of parameters in a model to decrease computational requirements and enhance efficiency in constrained environments. By removing redundant weights, neurons, or filters, pruning can also improve inference speed, energy efficiency, and, in some cases, model accuracy and robustness~\cite{cheng2024survey}. We explore following three types of neural network pruning: 

\textbf{1. Unstructured pruning:} Given a neural network with weights \( W = \{w_0, w_1, ..., w_K\} \), a dataset \( D = \{(x_i, y_i)\}_{i=1}^{N} \) consisting of input-output pairs \((x_i, y_i)\), and a target number of non-zero weights \( k \) (where \( k < K \)), unstructured pruning can be formulated as the following constrained optimization problem~\cite{cheng2024survey}:

\begin{equation}
\min_W L(W; D) = \min_W \frac{1}{N} \sum_{i=1}^{N} \ell(W; (x_i, y_i)), \quad \text{s.t.} \quad \|W\|_0 \leq k.
\end{equation}

In practice, for small to medium-sized models, unstructured pruning does not directly set weights to zero. Instead, it applies a binary mask \( M \) that determines which weights are active, setting the masked-out weights to zero~\cite{wang2020picking}. Typically, after pruning, the network undergoes retraining, either with fine-tuning or training from scratch, while keeping the mask \( M \) fixed. The pruned weights (those masked out) are not updated during this process.

\textbf{2. Structured pruning:} Structured pruning removes entire structural components of a neural network, such as channels, filters, neurons, or transformer attention heads, to achieve a targeted pruning ratio while minimizing performance degradation and maximizing speed improvements~\cite{he2018soft}. Structured pruning are hardware-friendly for deployment since they maximize speed improvement.

Formally, given a neural network with structural components \( S = \{s_1, s_2, ..., s_L\} \), where each \( s_i \) represents a set of channels, filters, neurons, or attention heads in layer \( i \), structured pruning searches for a pruned set \( S' = \{s'_1, s'_2, ..., s'_L\} \) such that:

\begin{equation}
    s'_i \subseteq s_i, \quad i \in \{1, ..., L\}
\end{equation}

\textbf{3. Global pruning:} Global pruning removes the least important parameters across the entire neural network rather than restricting pruning to individual layers or structures. This approach enables a more efficient allocation of remaining parameters by preserving the most critical weights regardless of their layer.

Formally, given a neural network with weights \( W = \{w_1, w_2, ..., w_K\} \) and a target number of nonzero weights \( k \) (where \( k < K \)), global pruning searches for a pruned weight set \( W' \subset W \) that satisfies:

\begin{equation}
    \| W' \|_0 \leq k, \quad \text{where} \quad W' = \{w \in W \mid \text{top-}k(|W|)\}
\end{equation}

Unlike structured pruning, which removes entire channels, filters, or neurons, global pruning removes individual weights across all layers based on their importance scores, often determined by magnitude, gradients, or other criteria~\cite{han2015learning, blalock2020state}. 

\textbf{Pre-train and post-train pruning:} Neural network pruning techniques we just discussed can be done either before training (pre-train)  or after training (post-train). Pre-train pruning prune neural networks at the very beginning before any training occurs to eliminate the computational cost of pre-training~\cite{cheng2024survey}. Let \( F(x; W_0 \odot M) \) represent a neural network, where \( W_0 \) denotes the initial, randomly sampled weights from a given initialization distribution, and \( M \) is a binary mask that determines which weights remain after pruning. Once the pruning is applied, the resulting sparse network \( F(x; W_0 \odot M) \) is trained directly from scratch. After \( t \) training epochs, the network converges to \( F(x; W_t \odot M) \), where \( W_t \) represents the modified weights. 

Post-train pruning is the most widely used pruning approach, where we first train a randomly initialized dense neural network \( F(x; W_0) \) until it converges to \( F(x; W_t) \). Then, we remove weights, filters, or neurons that contribute the least to the model’s performance. The resulting pruned model is represented as \( F(x; W'_t \odot M') \), where \( W'_t \) and \( M' \) denote the remaining weights and the corresponding pruning mask. This pruning step can be performed once (one-shot pruning), where all unimportant weights are pruned in a single step, multiple times (iterative pruning), where pruning is done gradually over several steps, and retraining (either fine-tune the pruned network \( F(x; W'_t \odot M') \) to restore performance, or train the remaining weights \( F(x; W_0 \odot M'') \) from scratch, where \( M'' \) is the final sparsity pattern after pruning.  

The lottery ticket hypothesis~\cite{frankle2018lottery} suggests that pruned models can achieve higher accuracy than their unpruned counterparts when retrained with their initial weights. Pruning has proven particularly beneficial for real-time applications like autonomous vehicles, where inference speed and energy efficiency are critical.

\subsection{Related Work}
Recent works have shown some relationship between pruning and post-hoc explanations. For example, Weber et al.~\cite{weber2023less} evaluate the impact of pruning on explainability. Using GradCAM~\cite{selvaraju2017grad} in VGG-16~\cite{simonyan2014very}, the authors explore the effects of varying compression rates on saliency maps. Their findings suggest that moderate pruning levels enhance explainability by reducing noise in saliency maps, while excessive pruning negatively affects attribution quality. However, their evaluation lacks a rigorous quantitative comparison of heatmaps and does not consider scenarios like pruning without fine-tuning or pre-train pruning. This leaves significant gaps in understanding how different pruning strategies influence explanation robustness.

Tan et a.~\cite{tan2024evaluating} investigate the effect of pruning on the robustness of explanations. The study prunes neurons identified as least important post-training, without fine-tuning, ensuring that model predictions remain unchanged. The results demonstrate that removing these neurons can lead to a collapse in explanations generated by XAI methods, even when predictions are unaffected. Khakzar et al.~\cite{khakzar2019improving} introduce an innovative approach where neural networks are pruned dynamically for individual inputs, retaining only neurons that significantly contribute to the prediction. While this approach highlights the potential of input-specific pruning, it may overlook the need for local importance when explaining individual samples. Additionally, the computational cost of input-specific pruning limits its scalability for large datasets and real-time applications. Abbasi et al.~\cite{abbasi2017interpreting} focus on simple pruning of filters in convolutional neural networks (CNNs) to improve interpretability. By selectively removing redundant filters, the approach simplifies network architecture, reducing noise in saliency maps and emphasizing critical features. Suwal et al.~\cite{suwal2025sparse} train ResNet-18 on ImageNette and compare post-hoc explanations from Vanilla Gradients (VG) and Integrated Gradients (IG) across pruning levels, evaluating sparsity and faithfulness.

While there is existing research on post-hoc explanation methods and training strategies independently, their combined impact on interpretability in vehicular AI remains underexplored. This study addresses this gap by systematically evaluating how training strategies (natural, adversarial, and pruning) affect explanation quality across different models and datasets relevant to vehicular applications. These insights contribute to designing secure and transparent AI systems for vehicles, balancing interpretability with performance.

\section{Motivation}

In this section, we provide the motivation behind using pruning as a technique for improving post-hoc explanations by discussing its impact on faithfulness and comprehensibility of explanations. 

A critical requirement for faithful explanations is that saliency maps must reliably reflect the model’s underlying decision-making process. In input-gradient-based methods such as Vanilla Gradient~\cite{simonyan2014deepinsideconvolutionalnetworks}, Integrated Gradients~\cite{sundararajan2017axiomaticattributiondeepnetworks}, and SmoothGrad~\cite{smilkov2017smoothgrad}, explanations are computed as the gradient of the model output with respect to input features:
\begin{equation}
    \mathcal{E}(\mathbf{x}) = \frac{\partial f(\mathbf{x})}{\partial \mathbf{x}}
\end{equation}
where, the explanation \( E(x) \) is computed as the gradient of the model's output with respect to the input features and \( f(x) \) represents the model's prediction for input \( x \). The gradient highlights how sensitive the model's prediction is to changes in each input feature.

These explanations are then visualized as saliency maps \( \mathcal{E}(\mathbf{x}) \) that highlight input features that contribute most to the model's decision. The faithfulness of these explanations hence depend on how well the gradient signal captures meaningful model behavior.

For naturally trained deep models, these gradients are often highly sensitive to small input perturbations, leading to noisy and unfaithful explanations. This is quantified by the gradient norm, which measures how much the model's output fluctuates with respect to input perturbations:

\begin{equation}
    \|\mathcal{E}(\mathbf{x})\|_2 = \left\| \frac{\partial f(\mathbf{x})}{\partial \mathbf{x}} \right\|_2
\end{equation}

If this norm is large, explanations are unstable and influenced by high-frequency artifacts rather than meaningful features. Prior work suggests that adversarial training reduces the gradient norm~\cite{gong2024structured}, thereby producing explanations that focus on more stable, robust features~\cite{etmann2019connection}.  

Pruning inherently reduces model complexity by removing unnecessary weights, which stabilizes gradients and mitigates noise, thus improving the clarity and faithfulness of explanations. This results in a smoother, lower-complexity function \( f(\mathbf{x}) \), which in turns leads to a more stable gradient distribution, ensuring that the saliency maps better capture true decision-making features. Mathematically, pruning results in:

\[
\|\mathcal{E}_{\text{pruned}}(\mathbf{x})\|_2 < \|\mathcal{E}_{\text{natural}}(\mathbf{x})\|_2
\]

Similarly, for an explanation to be useful, it must be sparse, focusing only on the most critical input features while ignoring irrelevant ones. This affects the comprehensibility of saliency maps. Pruning encourages weight sparsity in the network, which translates to fewer active neurons during inference. This acts as an implicit regularizer, enforcing an information bottleneck effect. As a result, pruned models tend to rely on a more compact and interpretable set of features, making their saliency maps naturally more comprehensible:

Therefore, we hypothesize that pruning significantly enhances interpretability by yielding saliency maps that are more stable, sparse, and representative of genuine model decisions, compared to natural or adversarial training alone.

\section{Experiment}\label{sec:methodology}
In this section, we describe the dataset, model, explanation methods and evaluation metrics used in our experiment. Our code is available at \url{https://github.com/sanishsuwal7/VehicleSecCopy/}. 

\subsection{Datasets}
\begin{itemize}

\item \textbf{LISA}: The LISA Traffic Sign Dataset~\cite{GTDLBenchICDCS} is a widely used benchmark for traffic sign detection and recognition in real-world driving environments. Collected from a vehicle-mounted camera in the United States, the dataset consists of 47 US traffic sign types with 7,855 annotations across 6,610 frames, capturing a wide range of real-world variations such as sign sizes (ranging from 6×6 to 167×168 pixels). Each traffic sign is carefully annotated with attributes such as sign type, position, size, occlusion status, and whether the sign is on a side road. 

\item \textbf{GTSRB:} The German Traffic Sign Recognition Benchmark (GTSRB)~\cite{stallkamp2012man} is another widely used dataset for traffic sign classification with more than 50,000 images of 43 different traffic sign classes, captured under real-world conditions with variations in lighting, occlusion, perspective distortion, and motion blur. Each image is annotated with the corresponding traffic sign label and bounding box, providing a comprehensive benchmark for evaluating deep learning models. 
\end{itemize}

\subsection{Model Architectures and Training}

\subsubsection{Natural training}\label{sec:natural training}

We use the VGG-16~\cite{simonyan2014very} architecture for LISA and GTSRB dataset. Both models are trained for 100 epochs with a batch size of 128. We optimized the network using Stochastic Gradient Descent (SGD) with a learning rate of 0.01, momentum of 0.9, and weight decay of 5e-4 to prevent overfitting. The loss was computed using a mean-reduction cross-entropy function. In addition, we trained a ResNet-18 model~\cite{he2016deep} pre-trained on ImageNet for LISA. The training was conducted for 100 epochs with a  batch size of 8, using the Ranger optimizer~\cite{Ranger} with a learning rate of 1e-4 and epsilon set to 1e-6 for stable convergence. We applied cross-entropy loss with mean reduction for classification. We discuss the results of ResNet model on LISA dataset in Appendix \ref{appendix:lisaresnet}. 

\subsubsection{Adversarial training}\label{sec:adv training}
To perform adversarial training, we generate adversarial examples that are produced from natural samples $\textbf{x} \in R^d$ by adding a perturbation vector $\delta \in R^d$. We use the PGD~\cite{madry2017towards} attack to obtain adversarial perturbations. The hyper-parameters of PGD attack in our adversarial training: $\epsilon$ = 0.01, attack step size = $\epsilon/10$, and number of iterations = 40. We also evaluate the quality of saliency maps when adversarial training strength ($\epsilon$) is increased to 0.1 for LISA dataset. Other training hyperparameters are kept as explained in Section \ref{sec:natural training}. 

\subsubsection{Neural Network Pruning}
We perform both \textbf{pre-} and \textbf{post-train} pruning on the models. While pre-train pruning involves pruning before training and optimizing a model to achieve a weight sparsity, post-train pruning is applied to a fully trained model. We evaluate models with and without fine-tuning following post-train pruning. Following are the types of pruning we evaluate in our work (see Section \ref{sec:pruning} for details): 
\begin{itemize}
    \item \textit{L1 unstructured pruning:} It focuses on removing individual weights from the model without regard for the structure of the neural network layers. We prune 20\% of weights in all convolutional layers and 10\% of weights in the output layers resulting in 19\% sparsity.
    \item \textit{Global pruning:} It selects and removes weights across the entire model, rather than within individual layers or based on specific layer criteria. We use the prune rate of 0.2 for the weights of all the layers achieving 20\% sparsity.
    \item \textit{Layered structured pruning:} It removes entire structures such as neurons, filters, or channels layer by layer. We use a rate of 0.1 for each layer except the last fully connected layer resulting in 10\% sparsity. We choose less sparsity limit for layered structured pruning because of high loss in accuracy compared to other pruning methods. 
\end{itemize} 

\textbf{Fine-tuning:} To improve model performance after pruning, we apply a fine-tuning strategy to optimize the pruned model while preserving its generalization capability. For both models, we initialize the unfine-tuned model, set up the optimizer similar to Section \ref{sec:natural training}, and retrain the model for 50 epochs.

\subsection{Explanation methods and metrics} 
For each model, we evaluate three widely used gradient-based explanations: Vanilla Gradient~\cite{simonyan2014deepinsideconvolutionalnetworks}, Integrated Gradients~\cite{sundararajan2017axiomaticattributiondeepnetworks}, and SmoothGrad~\cite{smilkov2017smoothgrad}) using sparsity~\cite{chalasani2020concise}, and faithfulness~\cite{rong22consistent}. Sparsity metric is computed as the difference between the given method's score and that of naturally trained model. Higher values are better in sparsity. Faithfulness is computed as the ROAD evaluation plot where sharper drop in accuracy represents faithful explanations. See details in Section \ref{sec:xai} and Section \ref{sec:metrics}. 

\begin{table}[]
\centering 
\caption{Model performance of natural training (Nat), adversarial training (Adv), unstructured L1 pruning (L1), global pruning (Global), and layered structured pruning (Layered) on VGG-based model for LISA dataset. Here, Pre-train, and Post-train means pruning before training and after training. FT means fine-tuning.}
\label{tab:lisa-vgg-performance}
\resizebox{0.45\textwidth}{!}{%
\begin{tabular}{@{}lccccc@{}}
\toprule
\textbf{Method/Model} & \textbf{Nat} & \textbf{Adv} & \textbf{L1} & \textbf{Global} & \textbf{Layered} \\ \midrule
\textbf{Pre-train} & 97.6 & 97.5 & 94.5 & 93.9 & 91.3 \\
\textbf{Post-train (FT)} & 97.6 & 97.5 & 95.3 & 95.4 & 94.9 \\
\textbf{Post-train (no FT)} & 97.6 & 97.5 & 94.8 & 94.7 & 93.8 \\ \bottomrule
\end{tabular}%
}
\end{table}

\begin{table}[]
\centering 
\caption{Model performance of natural training (Nat), adversarial training (Adv), unstructured L1 pruning (L1), global pruning (Global), and layered structured pruning (Layered) on VGG-based model for GTSRB dataset. Here, Pre-train, and Post-train means pruning before training and after training. FT means fine-tuning.}
\label{tab:gtsrbaccuracyadv1}
\resizebox{0.45\textwidth}{!}{%
\begin{tabular}{@{}lccccc@{}}
\toprule
\textbf{Method/Model} & \multicolumn{1}{l}{\textbf{Nat}} & \multicolumn{1}{l}{\textbf{Adv}} & \multicolumn{1}{l}{\textbf{L1}} & \multicolumn{1}{l}{\textbf{Global}} & \multicolumn{1}{l}{\textbf{Layered}} \\ \midrule
\textbf{Pre-train} & 99.9 & 99.2 & 98.1 & 99.5 & 94.4 \\
\textbf{Post-train (FT)} & 99.2 & 98.6 & 99.9 & 99.9 & 98.1 \\
\textbf{Post-train (no FT)} & 99.2 & 98.6 & 99.9 & 99.9 & 96.9 \\ \bottomrule
\end{tabular}%
}
\end{table}

\section{Results and Analysis}\label{sec:results}

\begin{figure}[h]
    \centering
    \includegraphics[width=0.62\linewidth]{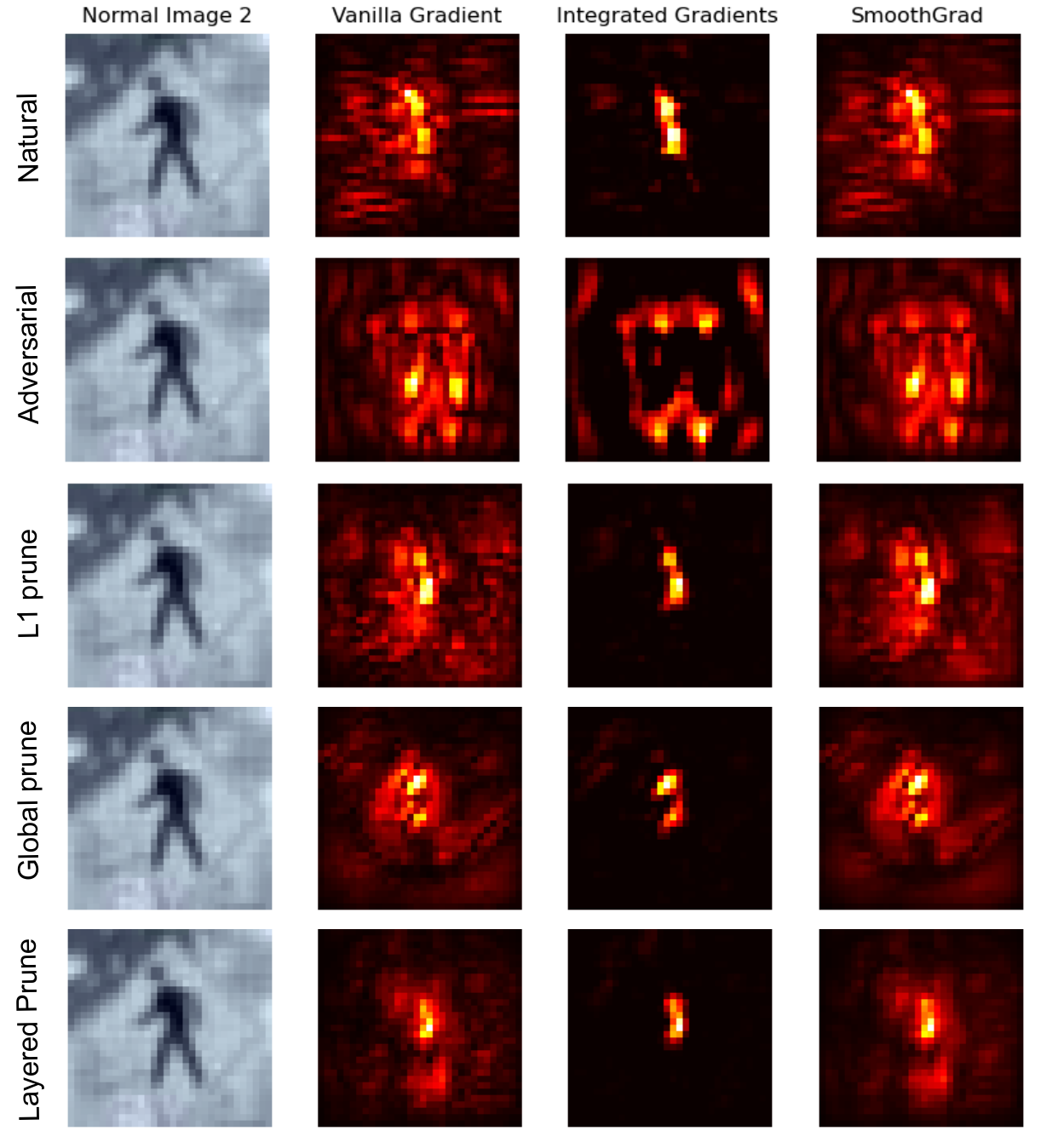}
    \caption{Saliency maps comparison for LISA dataset (VGG) between natural training, adversarial training and pre-train pruning.}
    \label{fig:vgg1}
\end{figure}

Table \ref{tab:lisa-vgg-performance} and \ref{tab:gtsrbaccuracyadv1} presents the classification accuracy of a VGG-based model trained on the LISA and GTSRB traffic sign dataset under natural training, adversarial training and different pruning strategies. As expected, naturally trained model has the highest accuracy compared to all other models. Among pruning strategies, fine-tuning after pruning is crucial for maintaining robust performance. Next, we evaluate the quality of saliency maps using different explanation methods in these models.

\subsection{Qualitative analysis}

\begin{figure}[h]
    \centering
    \includegraphics[width=0.65\linewidth]{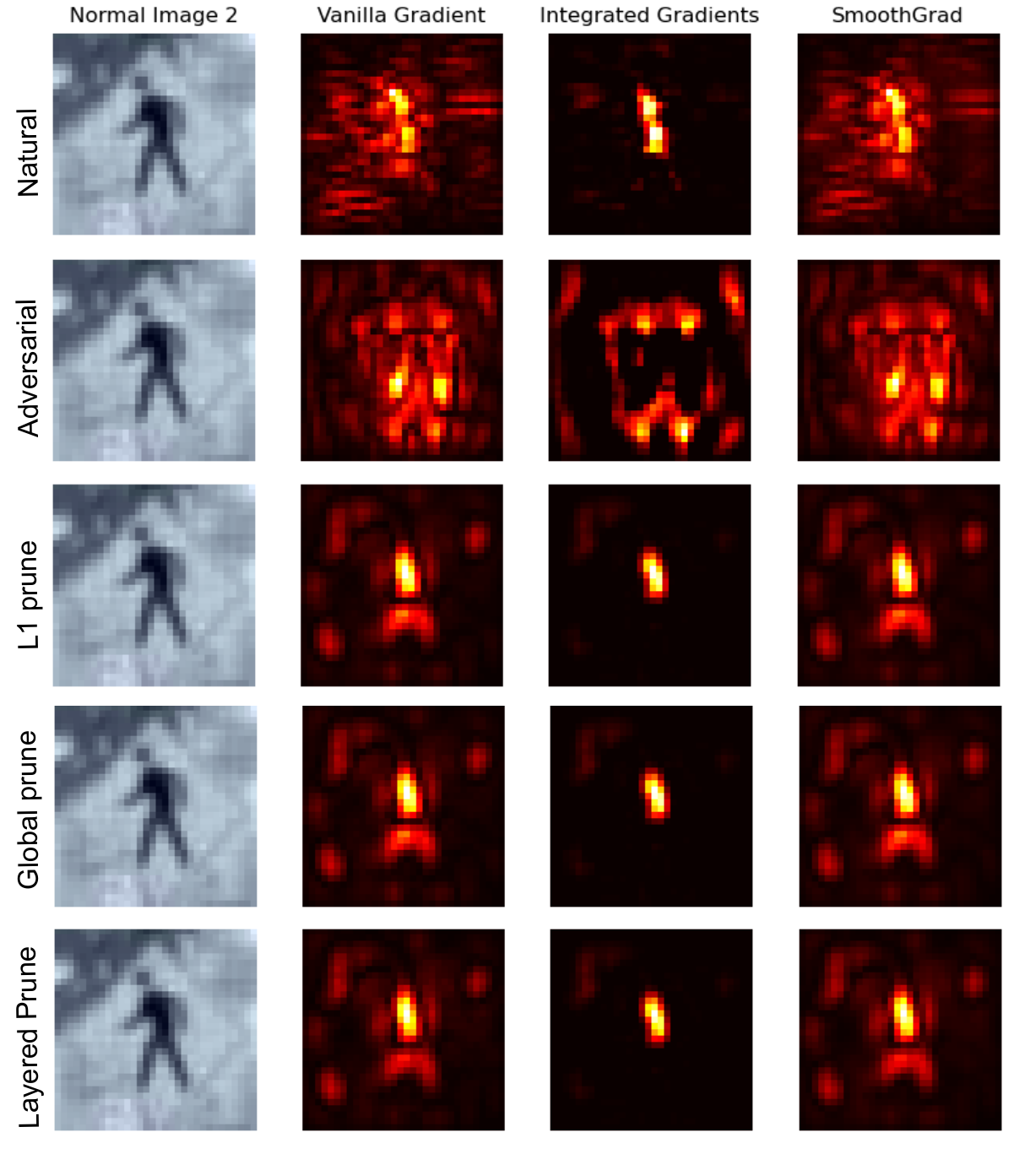}
      \caption{Saliency maps comparison for LISA dataset (VGG) between natural training, adversarial training and test-time pruning with no fine-tuning.}
    \label{fig:vgg2}
\end{figure}

\begin{figure}[h]
    \centering
    \includegraphics[width=0.65\linewidth]{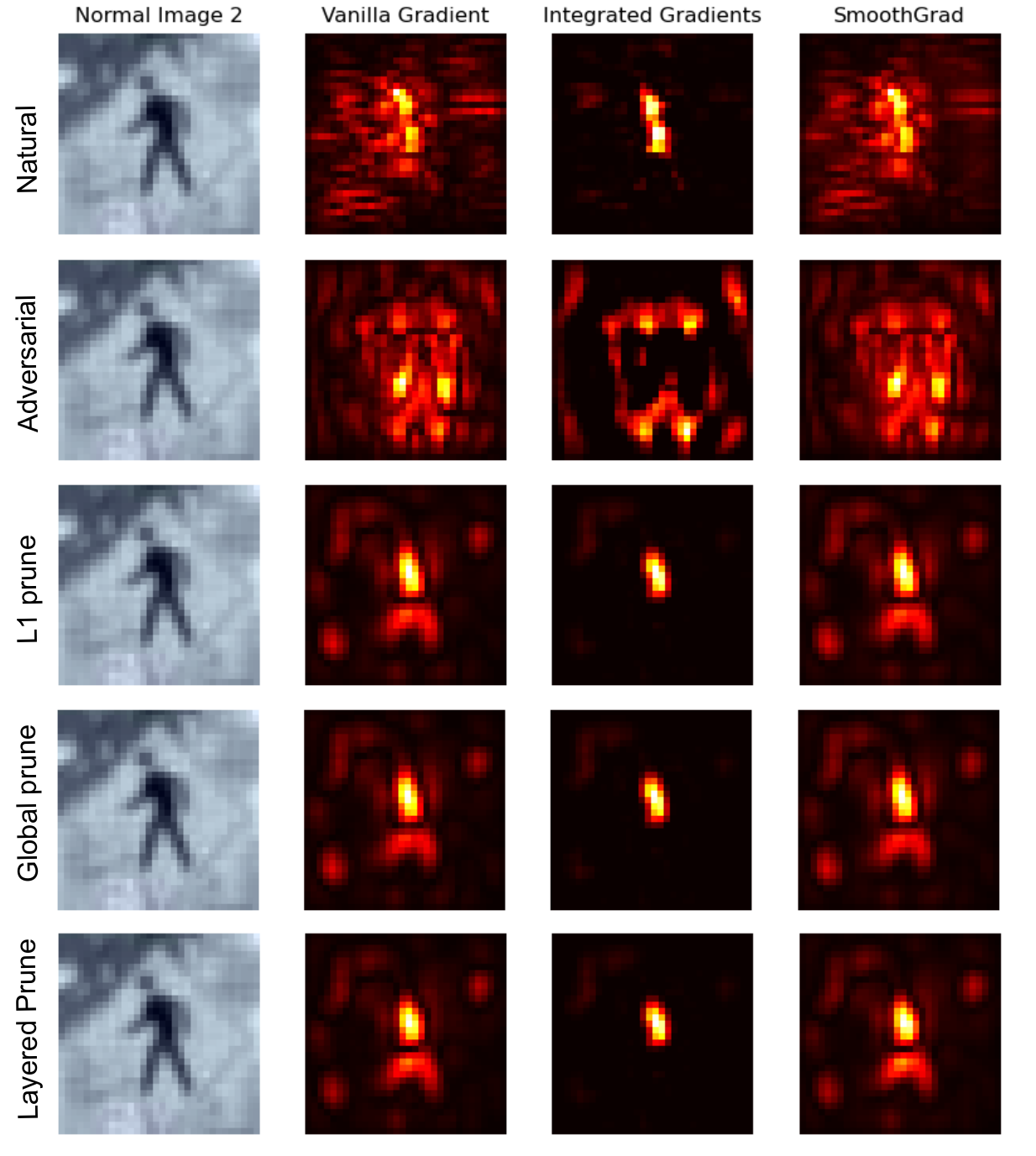}
        \caption{Saliency maps comparison for LISA dataset (VGG) between natural training, adversarial training and test-time pruning with fine-tuning.}
    \label{fig:vgg3}
\end{figure}

Figure \ref{fig:vgg1} shows that explanations from Vanilla Gradient in naturally trained models are noisy whereas saliency maps using Integrated Gradient are sparser by default. Since SmoothGrad is computed as an average of Vanilla Gradient explanations, the saliency maps are similar to the Vanilla Gradient but with an \textit{averaging} effect. Adversarially trained models produce more noisy saliency maps compared to the naturally trained models. While on larger datasets like ImageNet, adversarial training produces sparser saliency maps~\cite{chalasani2020concise}, this characteristic is not reciprocated with vehicle datasets that consists of low-dimension images. All pruned models, in pre-train pruning, has much sparser saliency maps than adversarially trained models. 

In Figure \ref{fig:vgg2}, we can observe that saliency maps for pruned models (without fine-tuning) focus on very relevant parts of the pedestrian image. Compared with saliency maps from naturally trained models and adversarially trained models, saliency maps from all three explanation methods, Vanilla Gradient, Integrated Gradient and SmoothGrad, focus on critical human features and are sparser and comprehensible. Similar saliency maps can be observed in Figure \ref{fig:vgg3} where saliency maps with pruned models are sparser and comprehensible. 

\begin{figure}[h]
    \centering
    \includegraphics[width=0.62\linewidth]{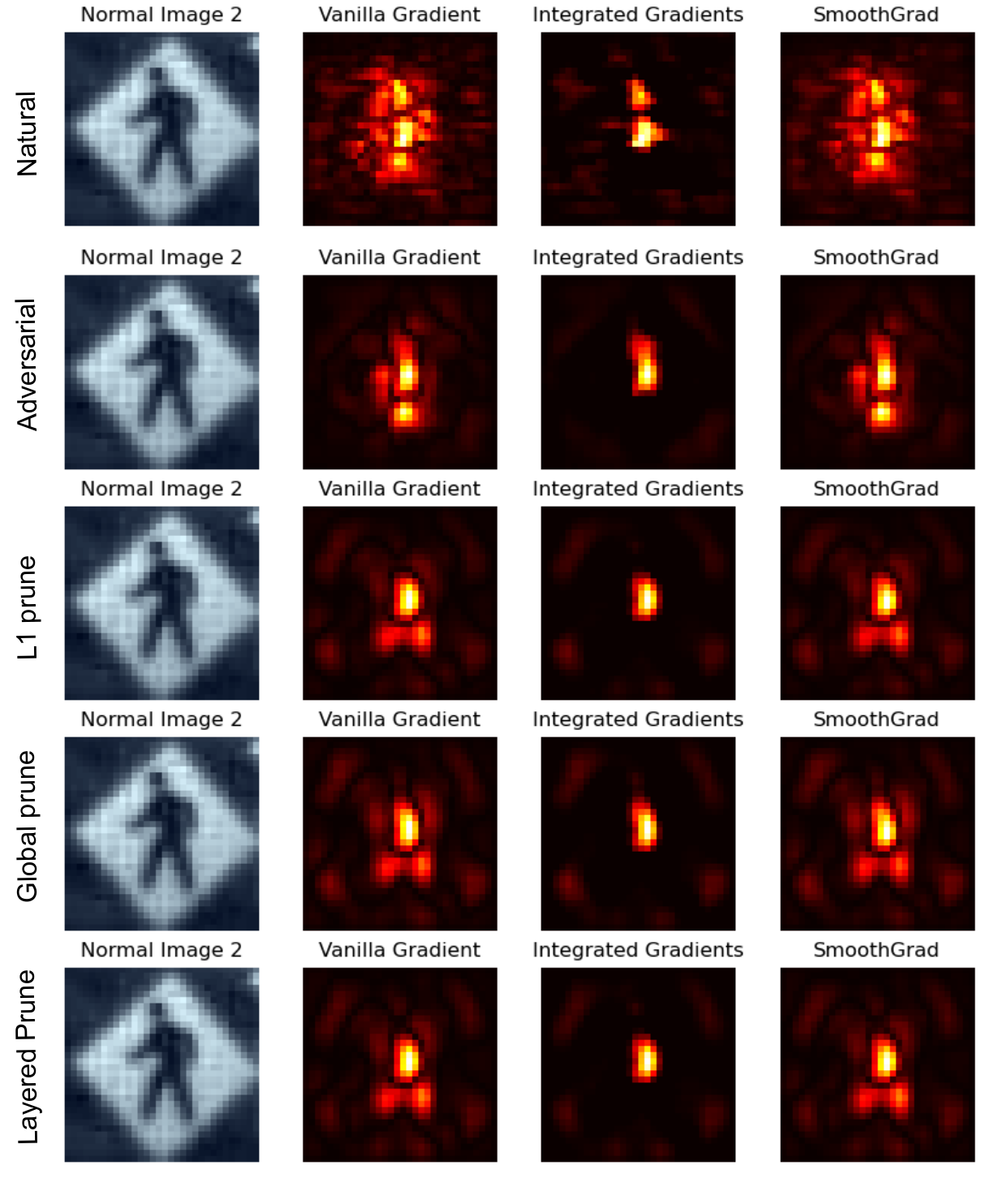}
    \caption{Saliency maps comparison for LISA dataset between natural training, adversarial training ($\epsilon=0.1$) and post-train pruning (with fine-tuning) for VGG.}
    \label{fig:mainvggepsilon2}
\end{figure}

However, if we increase the adversarial strength $\epsilon=0.1$, we get sparser saliency maps with adversarially trained model, as shown in Figure \ref{fig:mainvggepsilon2}. However, pruned models still seem to capture the specific human boundaries in the given image. Quantitatively, the benign accuracy of the model also decreases with increasing adversarial training robustness parameter (in our case, the model performance reduced from 97\% to 93\% accuracy), and pruned models also have much better faithfulness compared to adversarial trained models. See Section  \ref{sec:quantanalysis} for details.

\begin{figure}[h]
    \centering
    \includegraphics[width=0.62\linewidth]{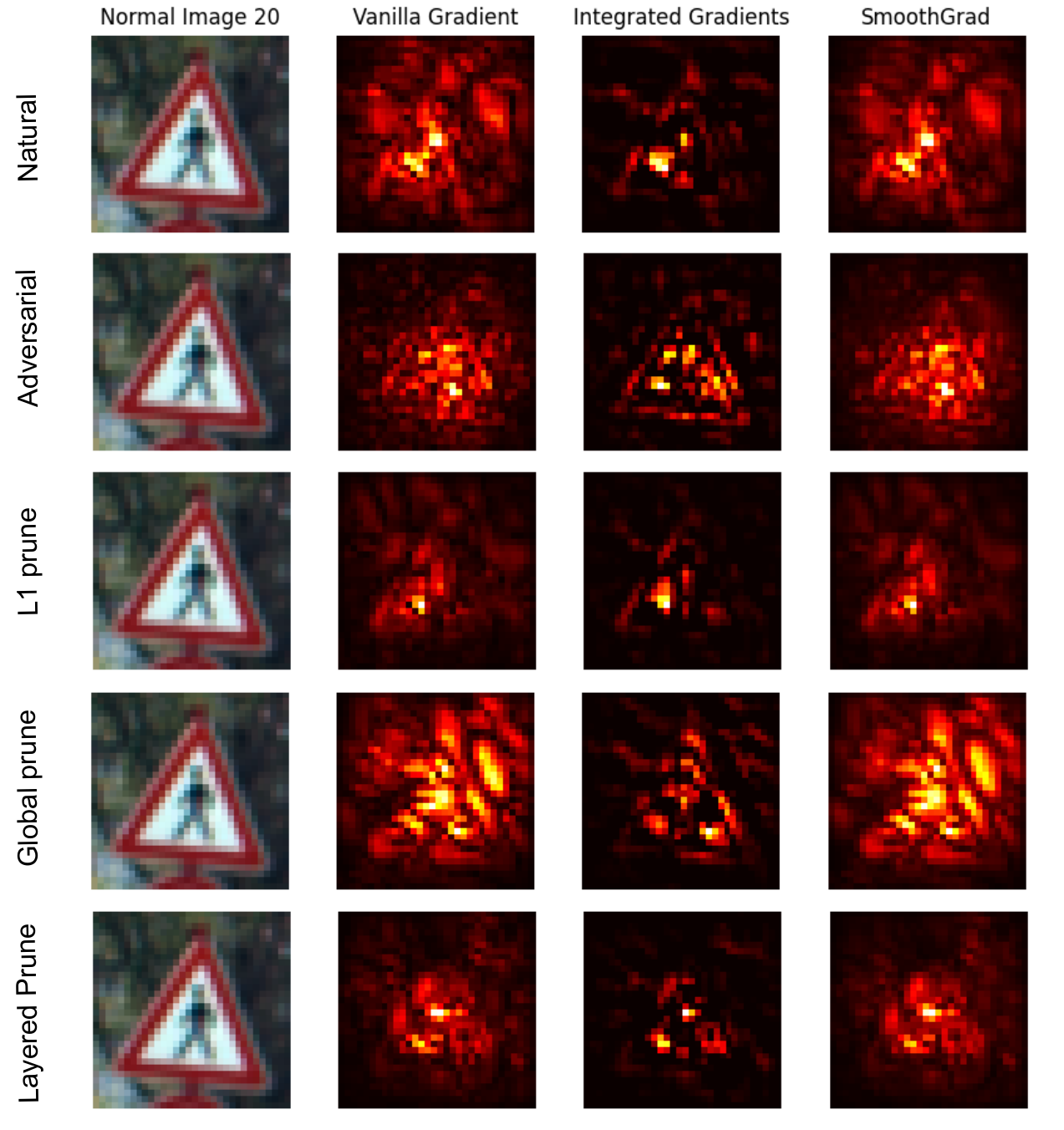}
    \caption{Saliency maps comparison for GTSRB dataset between natural training, adversarial training and pre-train pruning.}
    \label{fig:gtsrbadv1pretrain}
\end{figure}

Figure \ref{fig:gtsrbadv1pretrain} shows the saliency maps for GTSRB dataset where we can observe that saliency maps for naturally trained, adversarially trained and global pruned models using Vanilla Gradient are noisy whereas L1 prune and Layered Prune produces much clearer saliency maps. Similar to LISA, saliency maps using Integrated Gradient are sparser by default. However, all pruned models produce more comprehensible saliency maps. In Appendix \ref{appendix:gtsbradvtrain1posttrain}, we demonstrate saliency maps comparison for different training strategies and post-train pruning for GTSRB. In Appendix \ref{appendix:lisaresnet}, we demonstrate saliency maps for different training strategies on ResNet network for LISA dataset, revealing sparse and comprehensible saliency maps for pruned models.

\textbf{Discussion:} The increased clarity of saliency maps in pruned models suggest that pruning encourages feature selectivity by forcing the model to focus on the most critical input features for decision-making. This contrasts with naturally trained models, which, as observed, produce noisier saliency maps in vehicle datasets. One possible explanation is that pruning reduces model redundancy, eliminating unnecessary or redundant parameters that may contribute to spurious attributions in saliency maps. By contrast, natural training only aim to improve accuracy, relying on a broader range of features, thereby producing more noisy explanations.

\begin{table}[h]
\centering
\caption{Sparsity evaluation of Vanilla Gradient (VG), Integrated Gradient (IG) and SmoothGrad (SG) on different training strategy of LISA-VGG model. Here, adv means adversarial training and L1, Global and Layered means L1 unstructured pruning, Global pruning and layered structured pruning respectively. Pre-train and post-train means pruning before and after pruning and FT means fine tuning. Higher the better scores.}
\label{tab:sparsity-evaluation-vgg}
\resizebox{0.45\textwidth}{!}{%
\begin{tabular}{@{}llcccc@{}}
\toprule
\textbf{Method} & \textbf{Model} & \multicolumn{1}{l}{\textbf{Adv}} & \multicolumn{1}{l}{\textbf{L1}} & \multicolumn{1}{l}{\textbf{Global}} & \multicolumn{1}{l}{\textbf{Layered}} \\ \midrule
\textbf{VG} & \textbf{Pre-train} & -0.03 & -0.04 & -0.05 & -0.21 \\
 & \textbf{Post-train (FT)} & -0.03 & 0.00 & 0.00 & -0.01 \\
 & \textbf{Post-train (no FT)} & -0.03 & 0.00 & 0.00 & 0.00 \\ \hline
\textbf{IG} & \textbf{Pre-train} & -0.05 & -0.04 & -0.05 & -0.25 \\
 & \textbf{Post-train (FT)} & -0.05 & 0.00 & -0.01 & 0.00 \\
 & \textbf{Post-train (no FT)} & -0.05 & 0.00 & 0.00 & 0.00 \\ \hline
\textbf{SG} & \textbf{Pre-train} & -0.04 & -0.05 & -0.06 & -0.21 \\
 & \textbf{Post-train (FT)} & -0.04 & 0.00 & 0.01 & 0.00 \\
 & \textbf{Post-train (no FT)} & -0.04 & 0.00 & 0.00 & 0.00 \\ \bottomrule
\end{tabular}%
}
\end{table}

\subsection{Quantitative analysis}\label{sec:quantanalysis}

While qualitative analysis provides some visual cues of judgement, these are inherently bias to observers. Hence, we perform quantitative evaluation to judge the effectiveness of different training strategies for enhancing explanation quality.

\subsubsection{Sparsity of Explanations} 

Table \ref{tab:sparsity-evaluation-vgg} presents the sparsity scores of Vanilla Gradient (VG), Integrated Gradient (IG), and SmoothGrad (SG) on a VGG-based model trained with various pruning strategies for LISA dataset. The scores are measured using the Gini index and computed as the difference relative to naturally trained models, with higher values indicating better sparsity in attributions.

\begin{table}[]
\centering 
\caption{Sparsity evaluation of Vanilla Gradient (VG), Integrated Gradient (IG) and SmoothGrad (SG) on different training strategy of GTSRB model. Here, adv means adversarial training and L1, Global and Layered means L1 unstructured pruning, Global pruning and layered structured pruning respectively. Pre-train and post-train means pruning before and after pruning and FT means fine tuning. Higher the better scores.}
\label{tab:gtsrbsparsity1}
\resizebox{0.45\textwidth}{!}{%
\begin{tabular}{@{}llcccc@{}}
\toprule
\textbf{Method} & \textbf{Model} & \multicolumn{1}{l}{\textbf{Adv}} & \multicolumn{1}{l}{\textbf{L1}} & \multicolumn{1}{l}{\textbf{Global}} & \multicolumn{1}{l}{\textbf{Layered}} \\ \midrule
 & \textbf{Pre-train} & 0.03 & 0.00 & 0.00 & -0.02 \\
\textbf{VG} & \textbf{Post-train (FT)} & 0.03 & 0.00 & 0.00 & 0.00 \\
\textbf{} & \textbf{Post-train (no FT)} & 0.03 & 0.00 & 0.00 & 0.00 \\ \midrule
\textbf{} & \textbf{Pre-train} & 0.01 & 0.00 & 0.00 & -0.04 \\
\textbf{IG} & \textbf{Post-train (FT)} & 0.01 & 0.00 & 0.00 & -0.01 \\
\textbf{} & \textbf{Post-train (no FT)} & 0.01 & 0.00 & 0.00 & 0.00 \\ \midrule
\textbf{} & \textbf{Pre-train} & 0.03 & 0.00 & 0.00 & -0.02 \\
\textbf{SG} & \textbf{Post-train (FT)} & 0.03 & 0.00 & 0.00 & -0.01 \\
 & \textbf{Post-train (no FT)} & 0.03 & 0.00 & 0.00 & 0.00 \\ \bottomrule
\end{tabular}%
}
\end{table}

We can observe from Table \ref{tab:sparsity-evaluation-vgg} that adversarially trained models have consistently negative sparsity scores across all explanation methods, indicating that adversarial training does not improve sparsity and instead makes attributions more distributed compared to naturally trained models. This aligns with qualitative observations from Figures \ref{fig:vgg1}, \ref{fig:vgg2}, and \ref{fig:vgg3}, where adversarial training resulted in noisier saliency maps. However, we can also observe that pruned models do not necessarily have significantly positive sparsity scores; in fact, most pruning strategies have sparsity scores close to zero or slightly negative. This suggests that the overall distribution of feature importance remains similar between naturally trained and pruned models, with no drastic increase in sparsity as measured by the Gini index. However, this quantitative sparsity measurement does not fully capture the qualitative improvements observed in the saliency maps of pruned models. As shown in Figures \ref{fig:vgg2} and \ref{fig:vgg3}, the saliency maps from pruned models clearly highlight critical object boundaries, for pedestrian images. This indicates that while pruning does not necessarily lead to a mathematically sparser distribution of attributions, it refines feature selection, leading to qualitatively clearer and more human-comprehensible explanations.

In Table \ref{tab:vgg-epsion-2m}, we evaluate the sparsity for saliency maps between adversarially trained models ($\epsilon=0.1$) and post-train pruning with fine-tuning. We observed in Figure \ref{fig:mainvggepsilon2} that adversarially trained models produced sparser explanations when adversarial training strength was increased. However, similar to results in Table \ref{tab:sparsity-evaluation-vgg}, quantitatively, there is little to no gain in the sparsity scores. Even with increase in comprehensibility of saliency maps from adversarially trained models, these are still less sparse than all pruned models.

\begin{table}[]
\centering 
\caption{Sparsity evaluation of Vanilla Gradient (VG), Integrated Gradient (IG) and SmoothGrad (SG) on different training strategy of ResNet model. Here, adv means adversarial training ($\epsilon = 0.1$) and L1, Global and Layered means L1 unstructured pruning, Global pruning and layered structured pruning with fine-tuning. Higher the better scores.}
\label{tab:vgg-epsion-2m}
\resizebox{0.30\textwidth}{!}{%
\begin{tabular}{@{}ccccc@{}}
\toprule
\textbf{Method} & \textbf{Adv} & \textbf{L1} & \textbf{Global} & \textbf{Layered} \\ \midrule
VG & -0.01 & 0.00 & 0.00 & -0.01 \\
IG & -0.03 & 0.00 & -0.01 & 0.00 \\
SG & -0.01 & 0.00 & 0.01 & 0.00 \\ \bottomrule
\end{tabular}%
}
\end{table}

Similar results can be observed in Table \ref{tab:gtsrbsparsity1} for GTSRB dataset where adversarially trained models or pruned models do not necessarily have high sparsity. This confirms that these techniques maintain the overall distribution of feature importance but are able to focus on relevant parts of the image.

\textbf{Discussion:} These findings suggest that pruning does not simply increase attribution sparsity, but rather refines the relevance of explanations. It encourages the model to focus on essential object boundaries, leading to more interpretable saliency maps. This supports the hypothesis that pruning acts as an implicit regularizer, reducing redundant connections and forcing the network to make decisions based on a smaller set of critical features.

\begin{figure}[h]
    \centering
    \includegraphics[width=0.85\linewidth]{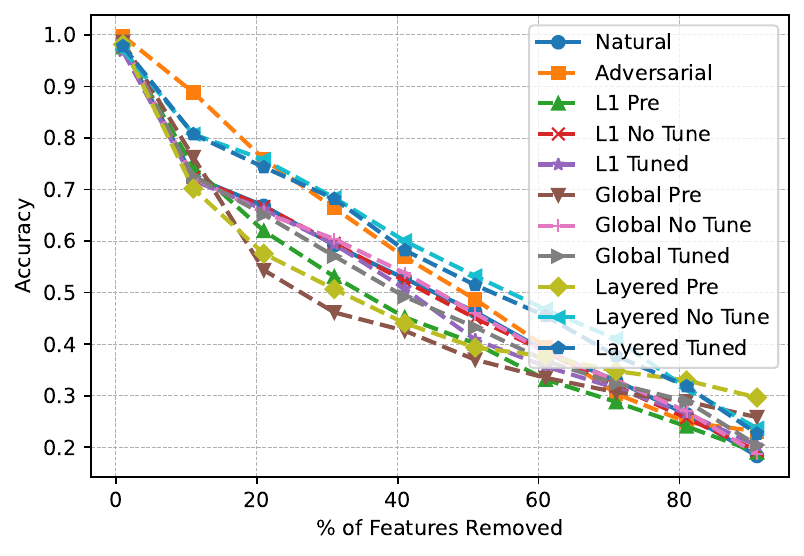}
    \caption{Faithfulness evaluation of Vanilla Gradient using ROAD on VGG for LISA using different strategies.}
    \label{fig:roadvggvanilla}
\end{figure}

\subsubsection{Faithfulness of Explanations}

\begin{figure}[h]
    \centering
    \includegraphics[width=0.80\linewidth]{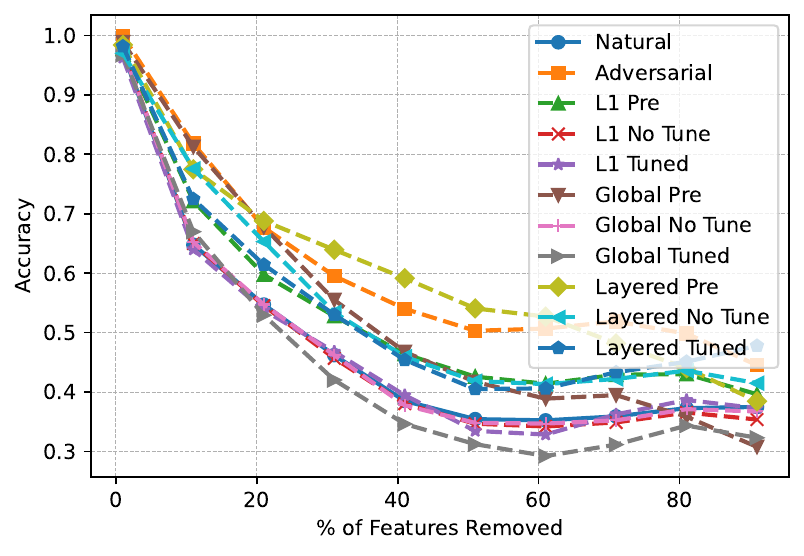}
    \caption{Faithfulness evaluation of Integrated Gradient using ROAD on VGG for LISA using different strategies.}
    \label{fig:roadigvanilla}
\end{figure}

Faithfulness measures whether the features highlighted by explanations accurately reflect the model's decision-making process. It is a critical metric, as effective explanations should help us understand how the model arrives at its predictions. We evaluate faithfulness using the ROAD (Remove and Debias) metric~\cite{rong22consistent}, where features are iteratively removed in order of decreasing importance (Most Relevant First (MoRF)), and the drop in model accuracy is observed. A sharper decline in accuracy indicates higher faithfulness, as it demonstrates that the removed features were indeed critical to the model's predictions.

\begin{figure}[h]
    \centering
    \includegraphics[width=0.80\linewidth]{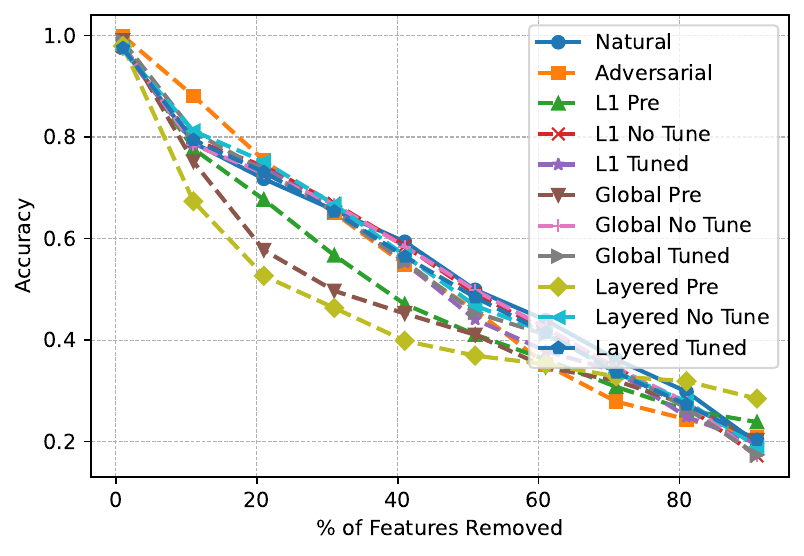}
    \caption{Faithfulness evaluation of SmoothGrad using ROAD on VGG for LISA using different strategies.}
    \label{fig:roadsgvanilla}
\end{figure}

Figure \ref{fig:roadvggvanilla} presents the faithfulness evaluation for Vanilla Gradient (VG) explanations across different training strategies. The results indicate that all pruned models (except Layered Pre-Prune) exhibit a steeper accuracy drop compared to naturally trained and adversarially trained models, signifying that pruning improves faithfulness. All pruning methods, in particular, leads to the most faithful explanations, as evidenced by a steeper accuracy decline within the 0–40\% feature removal range, meaning that the most relevant features identified by the explanations have a strong impact on model predictions. In contrast, naturally trained and adversarially trained models show a slower decay in accuracy, suggesting that their explanations do not strongly align with the model’s actual decision process, making them less faithful. These findings align with our earlier sparsity analyses, where we observed that naturally and adversarially trained models produced noisier explanations, making explanations less focused, contributing to their weaker faithfulness in the ROAD evaluation. Similar trends are observed in Figure \ref{fig:roadigvanilla}, which shows the faithfulness evaluation for Integrated Gradient (IG) explanations. The results confirm that pruning enhances explanation faithfulness, particularly for global fine-tuned pruning, which exhibits the steepest accuracy drop among all methods.

\begin{figure}[h]
    \centering
    \includegraphics[width=0.85\linewidth]{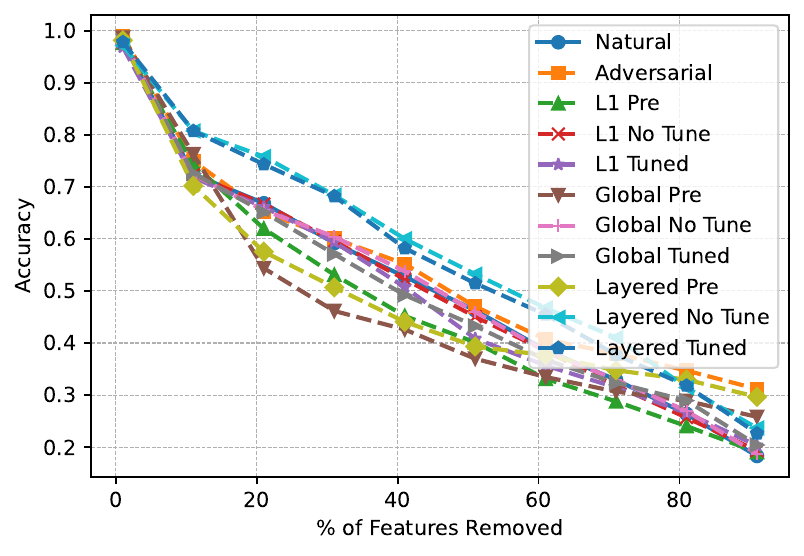}
    \caption{Faithfulness evaluation of Vanilla Gradient for LISA (adversarial training strength $\epsilon=0.1$)}
    \label{fig:roadvggeps2VG}
\end{figure}
\begin{figure}[h]
    \centering
    \includegraphics[width=0.85\linewidth]{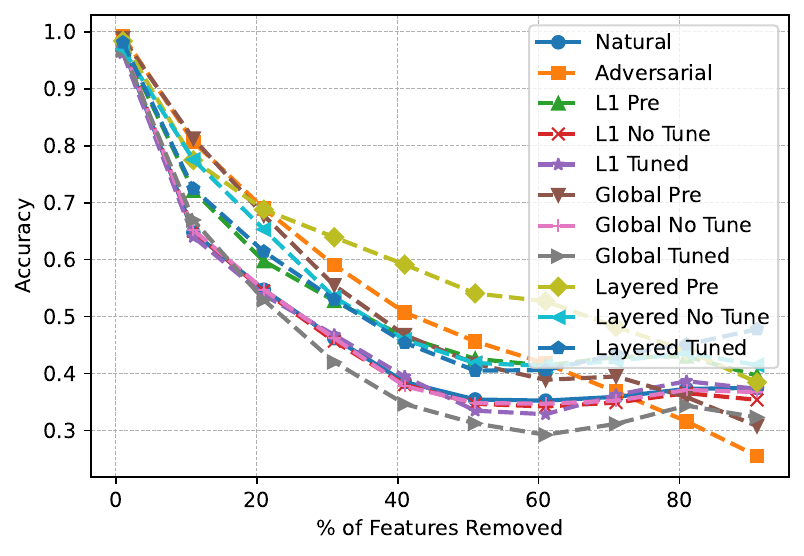}
    \caption{Faithfulness evaluation of Integrated Gradient LISA (adversarial training strength $\epsilon=0.1$)}
    \label{fig:roadvggeps2IG}
\end{figure}

Figure \ref{fig:roadsgvanilla} shows the faithfulness evaluation for SmoothGrad (SG) explanations, further validating that pruning enhances the faithfulness of saliency maps. The ROAD curves for Layered Pre-Prune, Global Pre-Prune, and L1 Pre-Prune (pre-train pruning methods) exhibit the steepest accuracy declines, demonstrating that pre-train pruning significantly improves the faithfulness of explanations. In Appendix \ref{appendix:lisaresnet}, we extend our faithfulness analysis to ResNet models, where we observe that the naturally trained ResNet model has an almost flat ROAD curve, signifying that the explanations generated from the naturally trained model are not faithful to the model’s actual decision process. This further supports our finding that pruning forces the model to rely on a subset of essential features, thereby improving explanation faithfulness and interpretability.

\begin{figure}[h]
    \centering
    \includegraphics[width=0.85\linewidth]{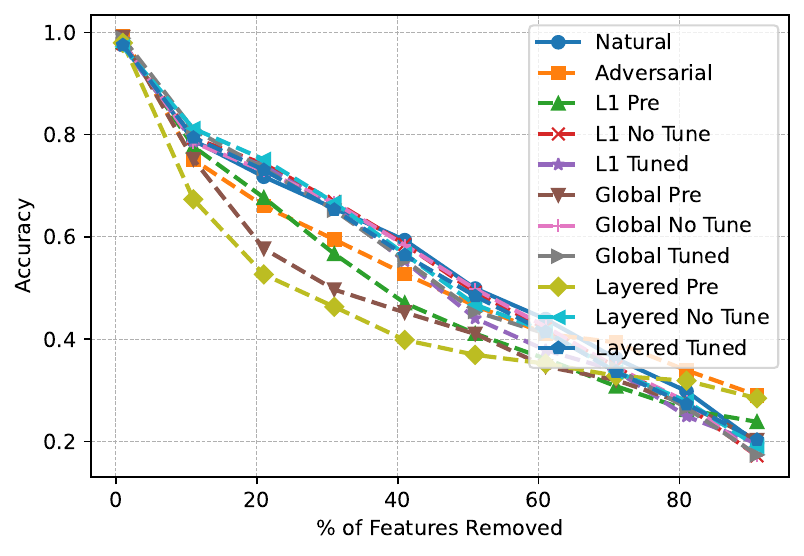}
    \caption{Faithfulness evaluation of Smooth Gradient LISA (adversarial training strength $\epsilon=0.1$)}
    \label{fig:roadvggeps2SG}
\end{figure}

As discussed before, increasing the adversarial training strength led to less noisy saliency maps. However, we demonstrated that this does not lead to any quantitative improvement in sparsity scores (see Table \ref{tab:vgg-epsion-2m}). Figures \ref{fig:roadvggeps2VG}, \ref{fig:roadvggeps2IG} and \ref{fig:roadvggeps2SG} demonstrate that pruning techniques, especially pre-train pruning, consistently emerge as the most effective strategies for generating faithful explanations, evident by the steepest drop in accuracy compared to adversarialy and naturally trained models. While there is an improvement over faithfulness explanations with adversarially trained model, pruning still generates the most faithful explanations.

\begin{figure}
    \centering
    \includegraphics[width=0.85\linewidth]{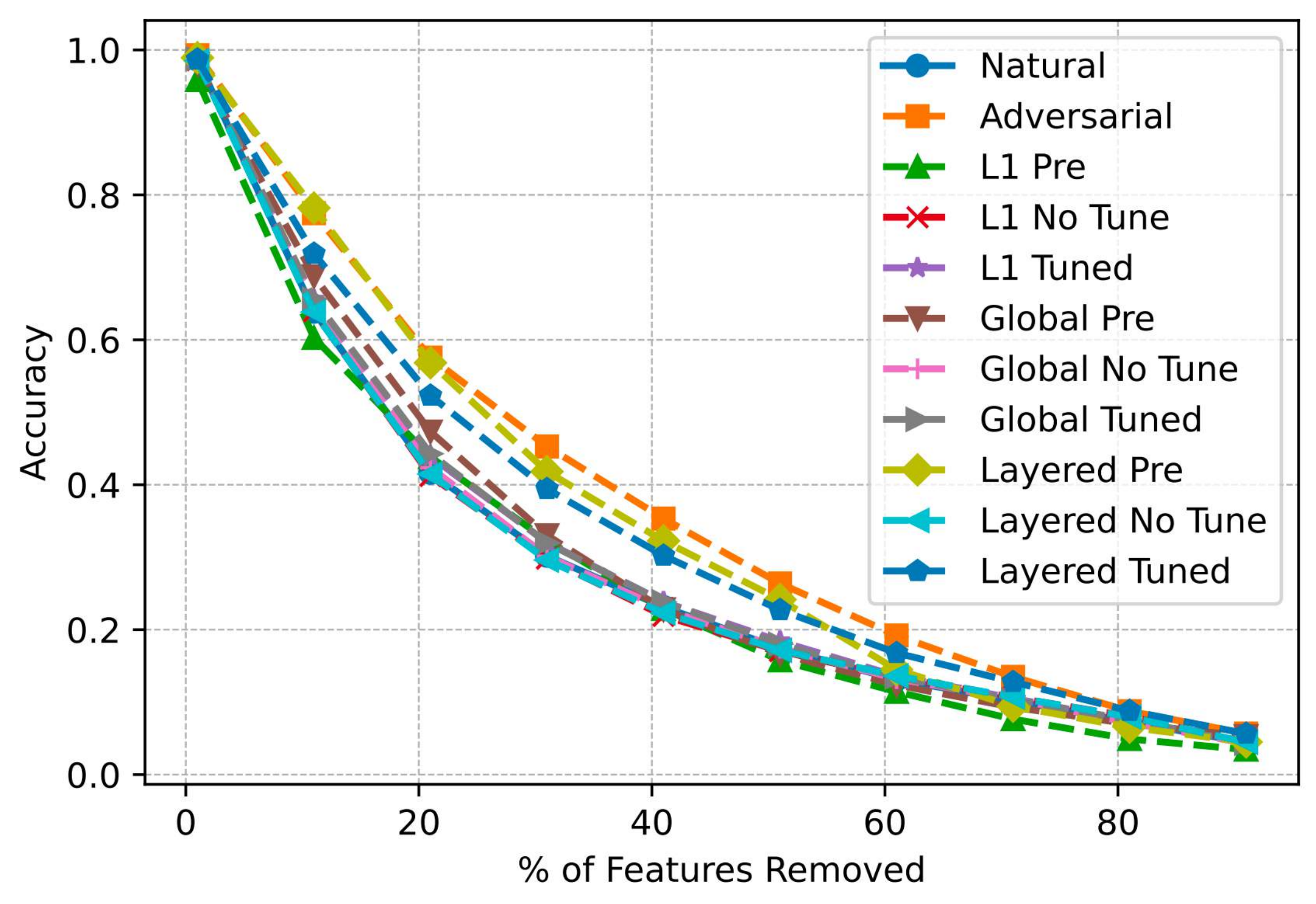}
    \caption{Faithfulness evaluation of Vanilla Gradient using ROAD for GTSRB using different strategies.}
    \label{fig:gtsrbroadvanillagrad}
\end{figure}

\begin{figure}[h]
    \centering
    \includegraphics[width=0.85\linewidth]{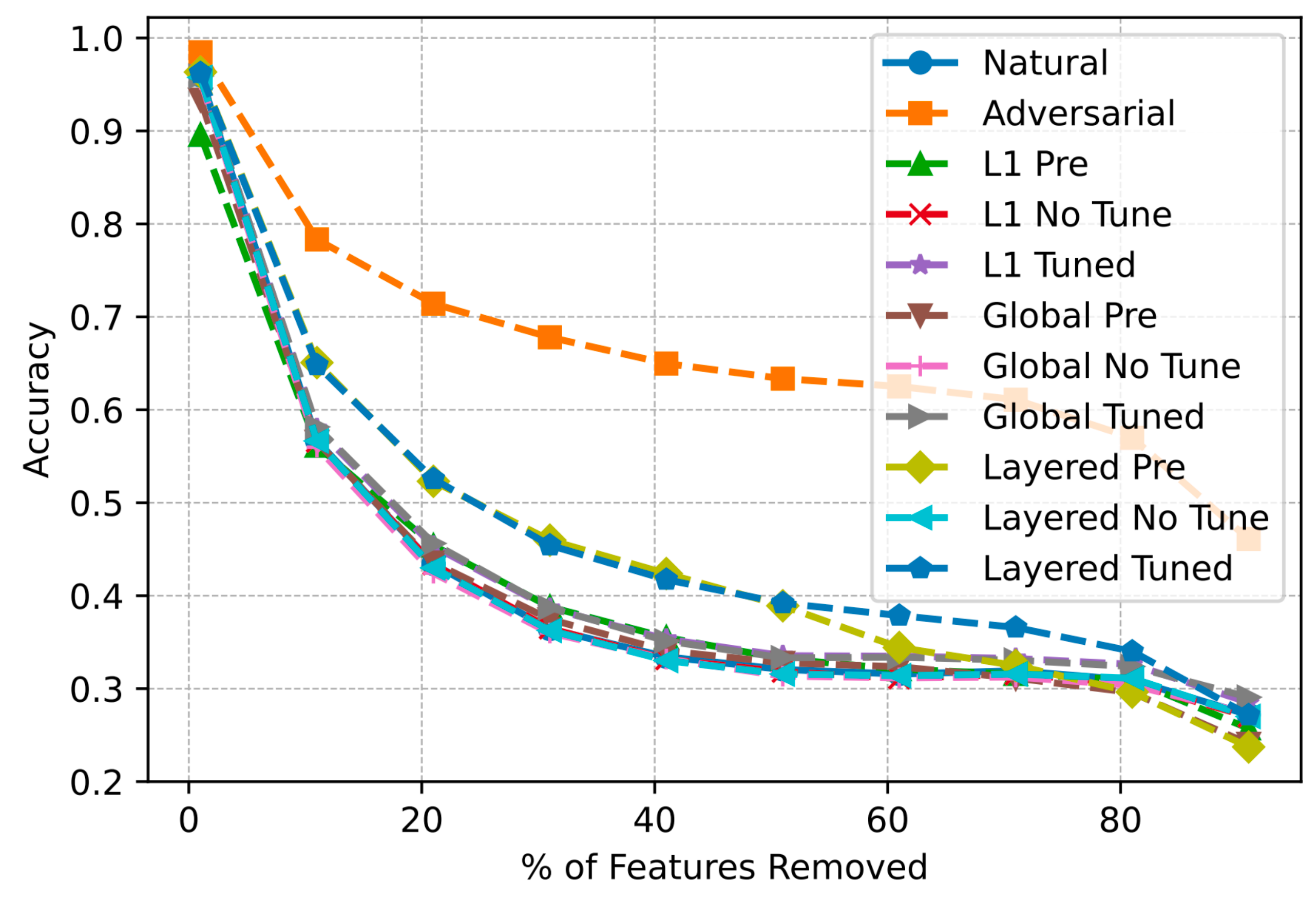}
        \caption{Faithfulness evaluation of Integrated Gradient using ROAD for GTSRB using different strategies.}
    \label{fig:gtsrbroadig}
\end{figure}

\begin{figure}[h]
    \centering
    \includegraphics[width=0.85\linewidth]{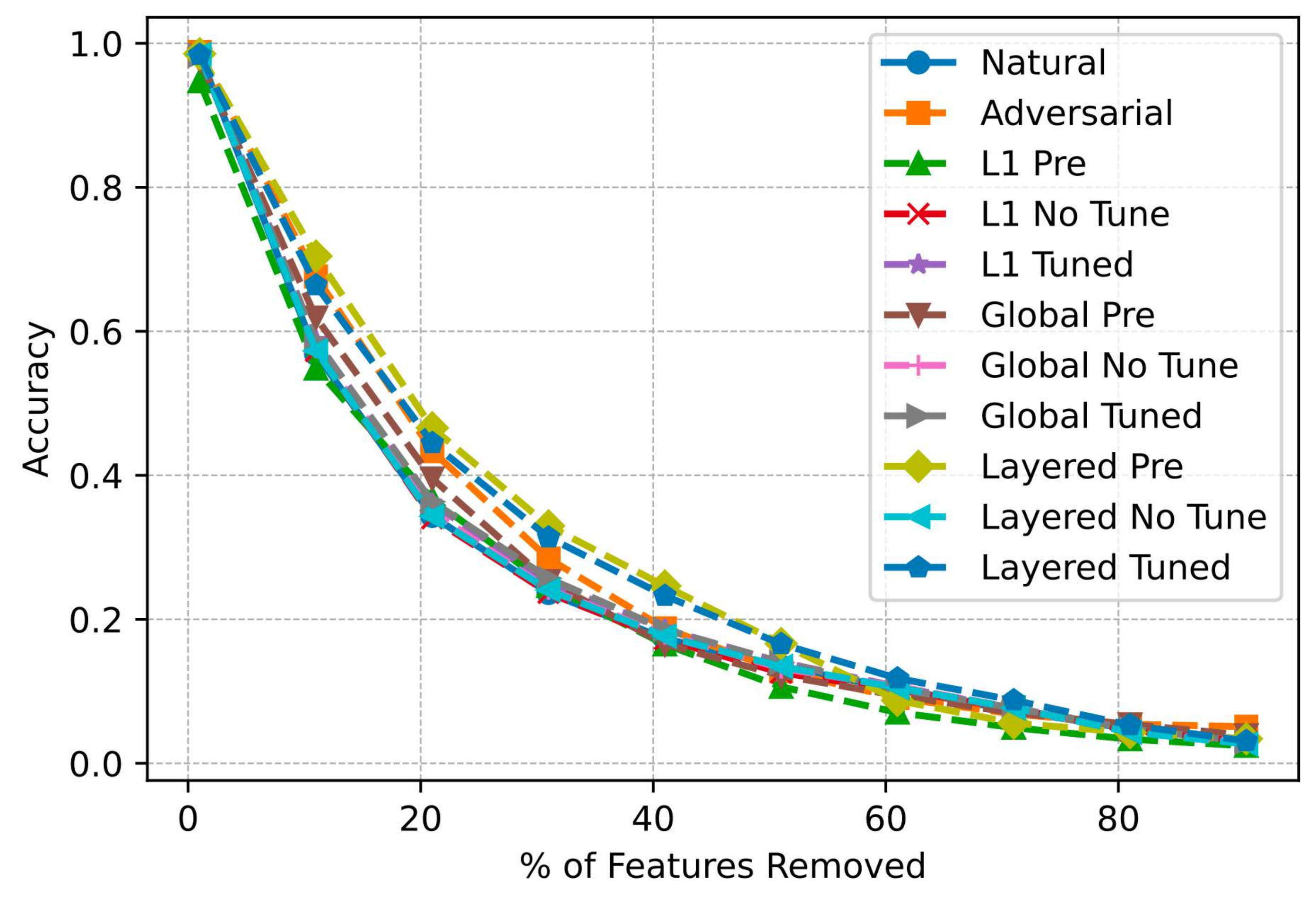}
           \caption{Faithfulness evaluation of SmoothGrad using ROAD for GTSRB using different strategies.}
    \label{fig:gtsrbroadsmoothgrad}
\end{figure}

These faithfulness results are consistent in GTSRB model. As shown in Figure \ref{fig:gtsrbroadvanillagrad}, in Vanilla Gradient, the adversarial models are the least faithful while the faithful measures in naturally trained models overlap with the pruned models. This is consistent with Integrated Gradient in Figure \ref{fig:gtsrbroadig}, and SmoothGrad in Figure \ref{fig:gtsrbroadsmoothgrad}.

\textbf{Discussion:} Pruning techniques consistently emerge as the most effective strategies for generating faithful explanations, indicating their ability of using the most important features in an input. This demonstrates that pruning not only reduces model complexity but also enhances the interpretability of AI systems. The faithfulness evaluation also underscores that explanations from naturally trained model are consistently less faithful meaning we cannot rely on post-hoc explanations of such models to understand a model prediction, and require training-strategy intervention for obtaining reliable explanations.

\section{Limitations}
We conducted our experiments on two popular traffic sign datasets: LISA and GTSRB and demonstrated how pruning a model can lead to comprehensible and faithful explanations improving model efficiency which is suitable for vehicular systems. However, while we explored different pruning techniques, the choice of the optimal pruning method for improving model transparency is dataset-dependent and largely empirical and task-dependent.

\section{Conclusion}\label{sec:conclusion}
This study underscores the critical role of training strategies in shaping the interpretability of AI systems for vehicular applications. By systematically evaluating the impact of natural training, adversarial training, and pruning strategies on the interpretability of deep learning models for traffic sign recognition, we assess how different training strategies affect sparsity, and faithfulness of explanations. Our findings demonstrate that pruning consistently enhances interpretability by generating comprehensible, and highly faithful saliency maps. While the suitable type of pruning is dataset-dependent and requires empirical validation, the results across two different datasets suggest that pruning plays a dual role, improving model efficiency while simultaneously enhancing explanation quality, and practitioners should prioritize pruning-based strategies to improve explanation quality.

\section*{Acknowledgment}
This work was supported by Toyota InfoTech Labs through Unrestricted Research Funds.

\bibliographystyle{plain}
\bibliography{reference}

\newpage 

\appendix

\section{LISA dataset - ResNet network}\label{appendix:lisaresnet}

Table \ref{tab:lisa-resnet-performance} shows the model performance on different training strategy with ResNet network on LISA dataset. We can clearly observe that adversarial training compromises benign performance significantly compared to the pruned models.

\begin{table}[]
\centering 
\caption{Model performance of natural training (Nat), adversarial training (Adv), unstructured L1 pruning (L1), global pruning (Global), and layered structured pruning (Layered) on ResNet-based model for LISA dataset. Here, Pre-train, and Post-train means pruning before training and after training. FT means fine-tuning.}
\label{tab:lisa-resnet-performance}
\resizebox{0.45\textwidth}{!}{%
\begin{tabular}{@{}lccccc@{}}
\toprule
\textbf{Model} & \textbf{Nat} & \textbf{Adv} & \textbf{L1} & \textbf{Global} & \textbf{Layered} \\ \midrule
Pre-train & 98.3 & 84.2 & 97.1 & 96.1 & 95.1 \\
Post-train (FT) & 98.3 & 84.2 & 96.8 & 98.8 & 99.4 \\
Post-train (no FT) & 98.3 & 84.2 & 97.4 & 97.9 & 89.1 \\ \bottomrule
\end{tabular}%
}
\end{table}

\begin{figure}[h]
    \centering
    \includegraphics[width=0.68\linewidth]{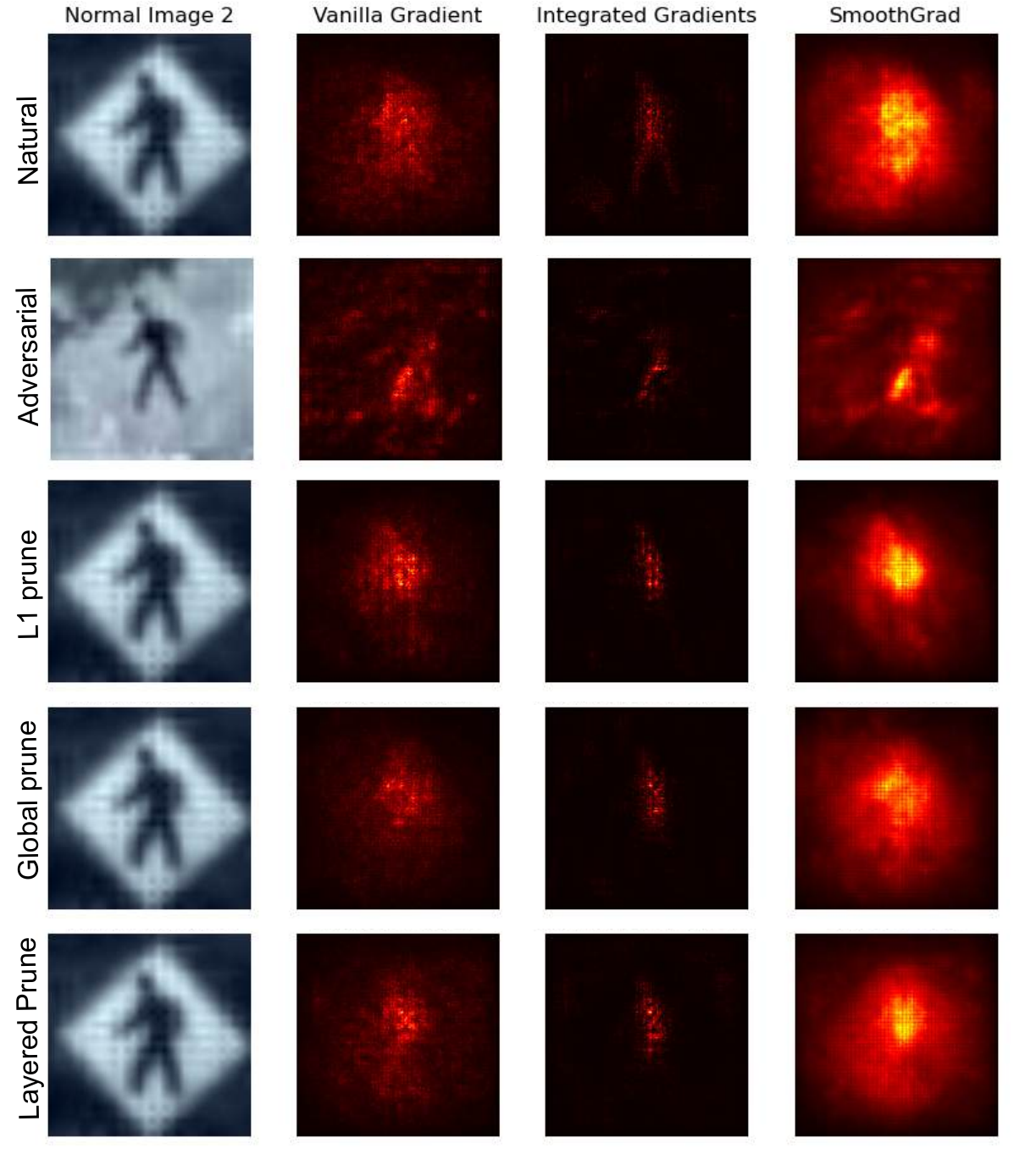}
    \caption{Saliency maps comparison for LISA dataset between natural training, adversarial training and pre-train pruning for ResNet model.}
    \label{fig:resnet1}
\end{figure}

\begin{figure}[h]
    \centering
    \includegraphics[width=0.68\linewidth]{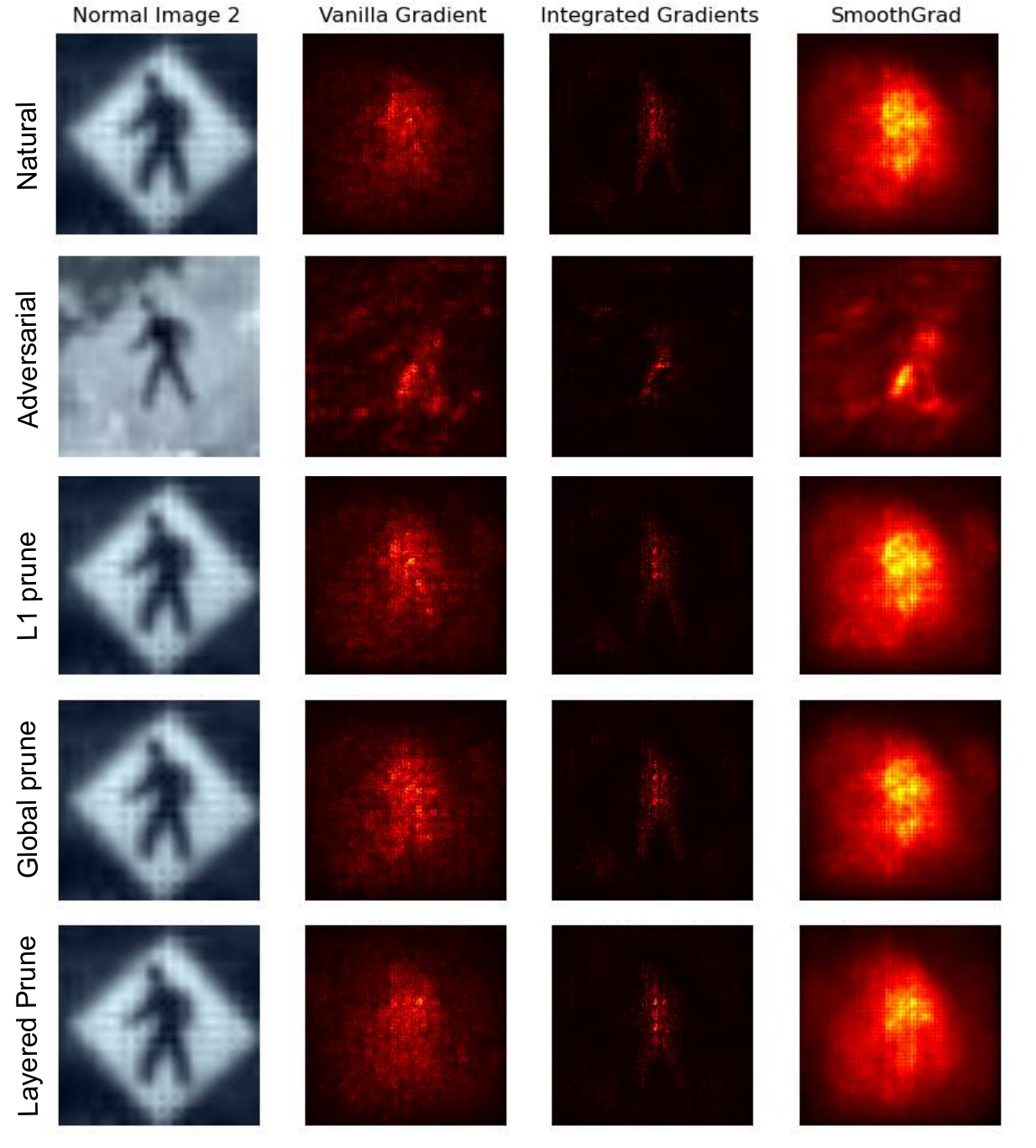}
      \caption{Saliency maps comparison for LISA dataset between natural training, adversarial training and test-time pruning with no fine-tuning for ResNet model.}
    \label{fig:resnet2}
\end{figure}

\begin{figure}[h]
    \centering
    \includegraphics[width=0.68\linewidth]{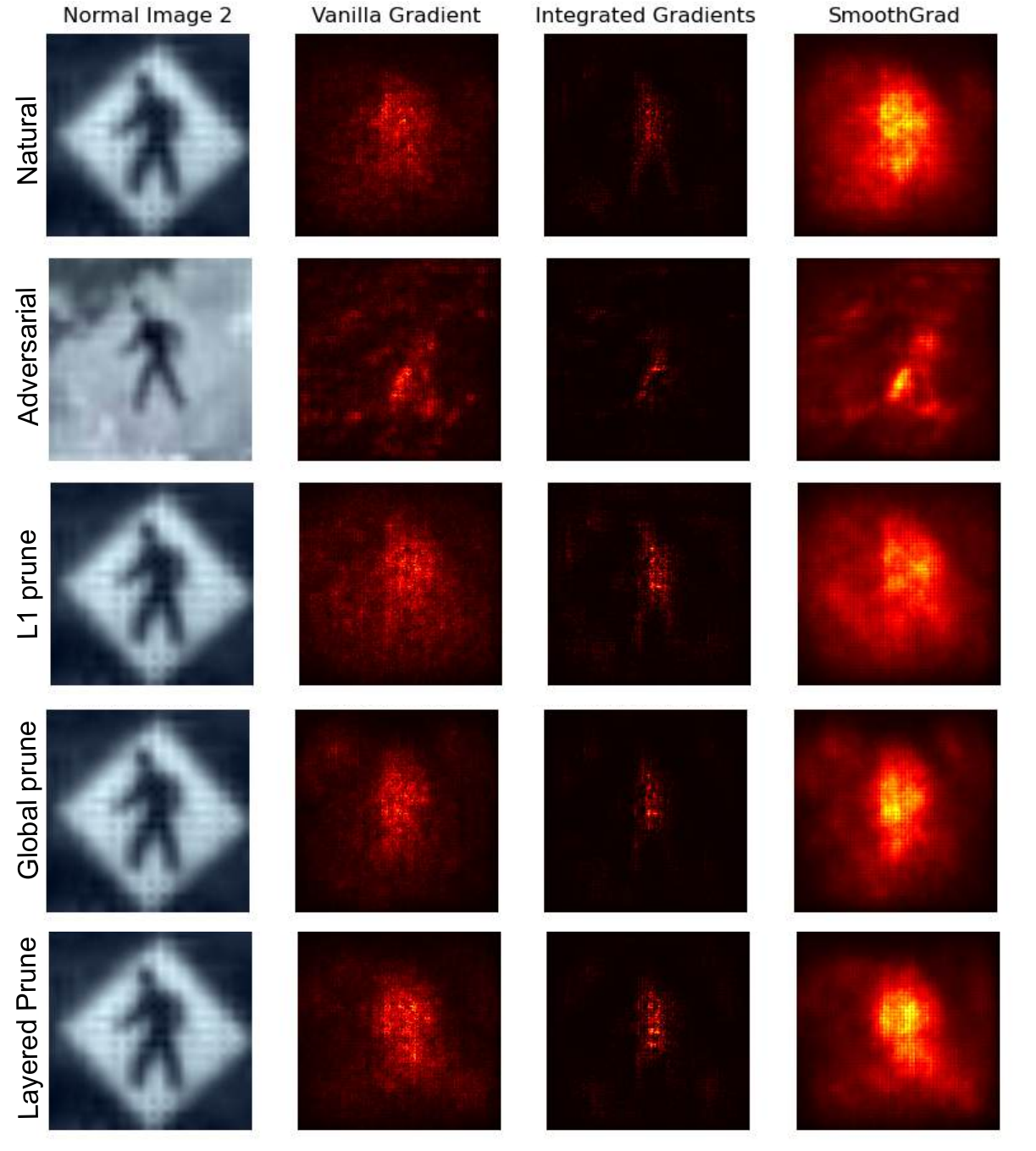}
        \caption{Saliency maps comparison for LISA dataset between natural training, adversarial training and test-time pruning with fine-tuning for ResNet model.}
    \label{fig:resnet3}
\end{figure}

Figures \ref{fig:resnet1}, \ref{fig:resnet2} and \ref{fig:resnet3} show the saliency maps for different training strategies on ResNet network, revealing sparse and comprehensible saliency maps for pruned models. As before, Integrated Gradients create much cleaner saliency maps. However, in Vanilla Gradient and SmoothGrad, we can observe less noisy saliency maps with pruned models.

Table \ref{tab:sparsity-evaluation-resnet} presents a comparative evaluation of sparsity scores. As before, there is no significant gain in sparsity with pruned models. However, the noisy saliency maps in adversarially trained models are validated by negative scores in the table.

\begin{table}[h]
\centering 
\caption{Sparsity evaluation of Vanilla Gradient (VG), Integrated Gradient (IG) and SmoothGrad (SG) on different training strategy of ResNet model for LISA dataset. Here, adv means adversarial training and L1, Global and Layered means L1 unstructured pruning, Global pruning and layered structured pruning. Pre-train, and Post-train means pruning before training and after training. FT means fine-tuning. Higher the better scores.}
\label{tab:sparsity-evaluation-resnet}
\resizebox{0.45\textwidth}{!}{%
\begin{tabular}{@{}llcccc@{}}
\toprule
\multicolumn{1}{c}{\textbf{Method}} & \multicolumn{1}{c}{\textbf{Model}} & \textbf{Adv} & \textbf{L1} & \textbf{Global} & \textbf{Layered} \\ \midrule
\textbf{VG} & Pre-train & -0.03 & 0.01 & 0.00 & -0.01 \\
 & Post-train (FT) & -0.03 & 0.00 & 0.00 & -0.02 \\
 & Post-train (no FT) & -0.03 & -0.01 & 0.02 & 0.00 \\ \hline
\textbf{IG} & Pre-train & -0.01 & 0.01 & 0.00 & 0.00 \\
 & Post-train (FT) & -0.01 & 0.00 & 0.02 & 0.00 \\
 & Post-train (no FT) & -0.01 & 0.00 & 0.00 & -0.01 \\ \hline
\textbf{SG} & Pre-train & -0.06 & 0.02 & 0.00 & -0.01 \\
 & Post-train (FT) & -0.06 & -0.01 & 0.03 & 0.00 \\
 & Post-train (no FT) & -0.06 & 0.00 & 0.00 & -0.02 \\ \bottomrule
\end{tabular}%
}
\end{table}

\begin{figure}[h]
    \centering
    \includegraphics[width=0.85\linewidth]{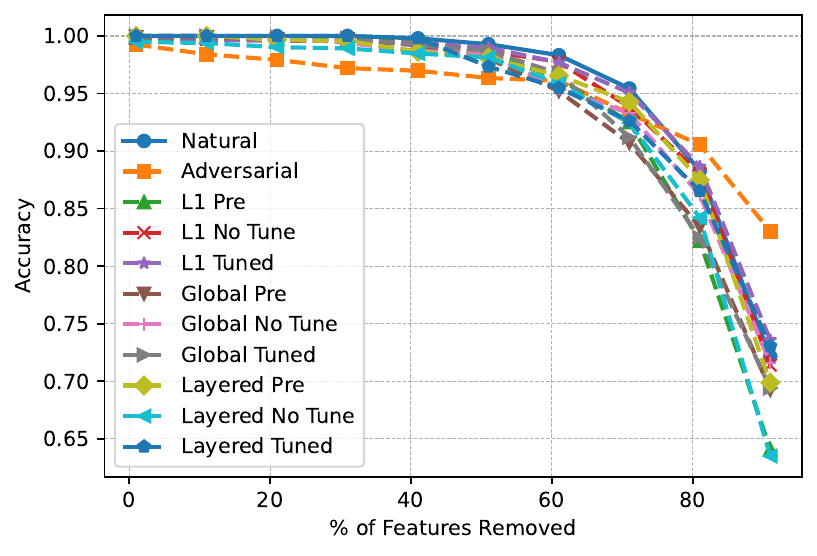}
    \caption{Faithfulness evaluation of Vanilla Gradient using ROAD on ResNET using different strategies.}
    \label{fig:roadRESNETvanilla}
\end{figure}

Figure \ref{fig:roadRESNETvanilla} shows the faithfulness evaluation for Vanilla Gradient method, where adversarial trained model has sharper drop in accuracy compared to other approaches especially up-to feature removal of 50\%. All other pruning approaches however surpass adversarial training method following this. As seen in the plot, the naturally trained model has almost a flat curve, signifying that the explanations produced from naturally trained model are not faithful to the model. This raises questions on the the quality of saliency maps using Vanilla Gradient (VG) with the naturally trained model as they do not truly reflect the underlying model. 

\begin{figure}[h]
    \centering
    \includegraphics[width=0.85\linewidth]{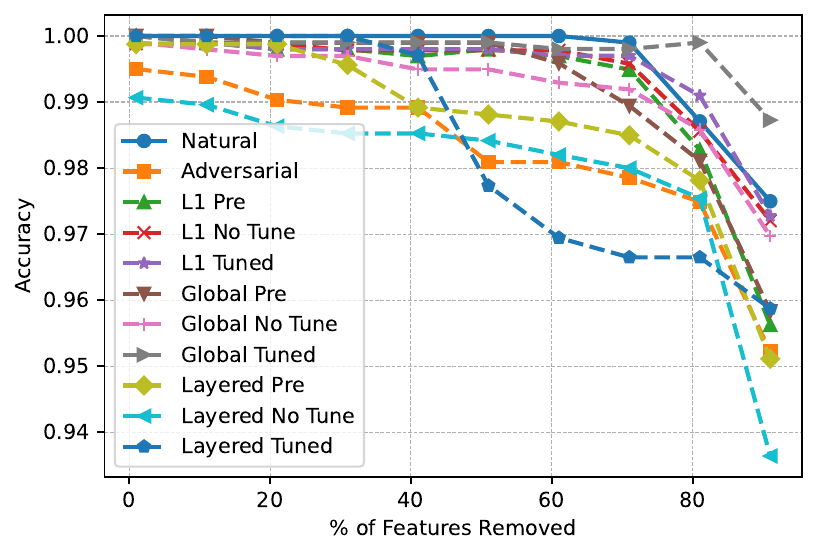}
    \caption{Faithfulness evaluation of Integrated Gradient using ROAD on ResNet using different strategies.}
    \label{fig:roadRESNETig}
\end{figure}

Similar observations can be made from Figure \ref{fig:roadRESNETig} that shows the faithfulness evaluation for Integrated Gradient method. The results demonstrate that pruning generally enhances explanation faithfulness, as indicated by a sharper drop in accuracy when the most relevant features are removed. This is prominent with layered (no fine tune method). As more features are removed, layered (with fine tune) shows the sharpest decline in accuracy, confirming its faithfulness. In comparison, naturally trained model has a flat line, confirming that natural training does not lead to faithful explanations.

\begin{figure}[h]
    \centering
    \includegraphics[width=0.85\linewidth]{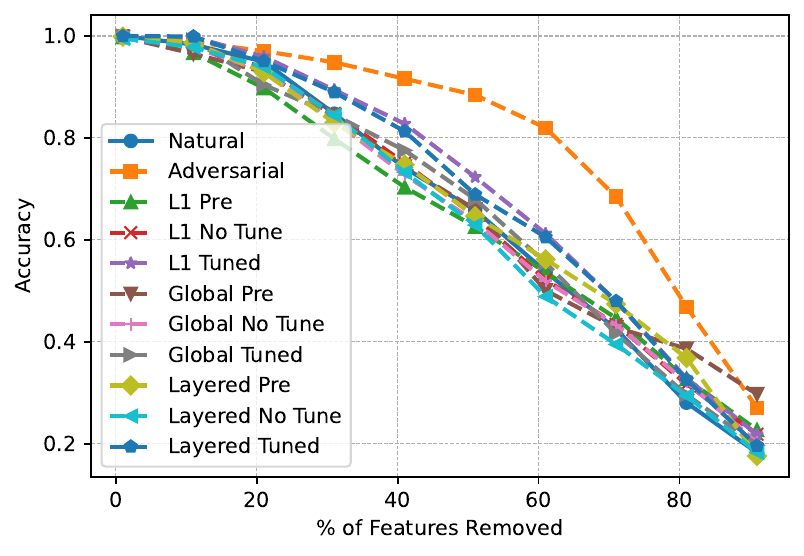}
    \caption{Faithfulness evaluation of Smooth Gradient using ROAD on ResNet using different strategies.}
    \label{fig:roadresnetSG}
\end{figure}

Figure \ref{fig:roadresnetSG}  that shows the faithfulness evaluation for SmoothGrad further validates that pruning enhances the faithfulness of saliency maps, as models trained with pruning generally exhibit a steeper accuracy decline when key features are removed. As observed in the figure, Layered Pre, L1 Pre and Global Pre (pre-train pruning) have the steepest decline demonstrating that pruning significantly improves the faithfulness of explanations.

\section{GTSRB Dataset: Post-Train Prune}\label{appendix:gtsbradvtrain1posttrain}

\begin{figure}[h]
    \centering
    \includegraphics[width=0.85\linewidth]{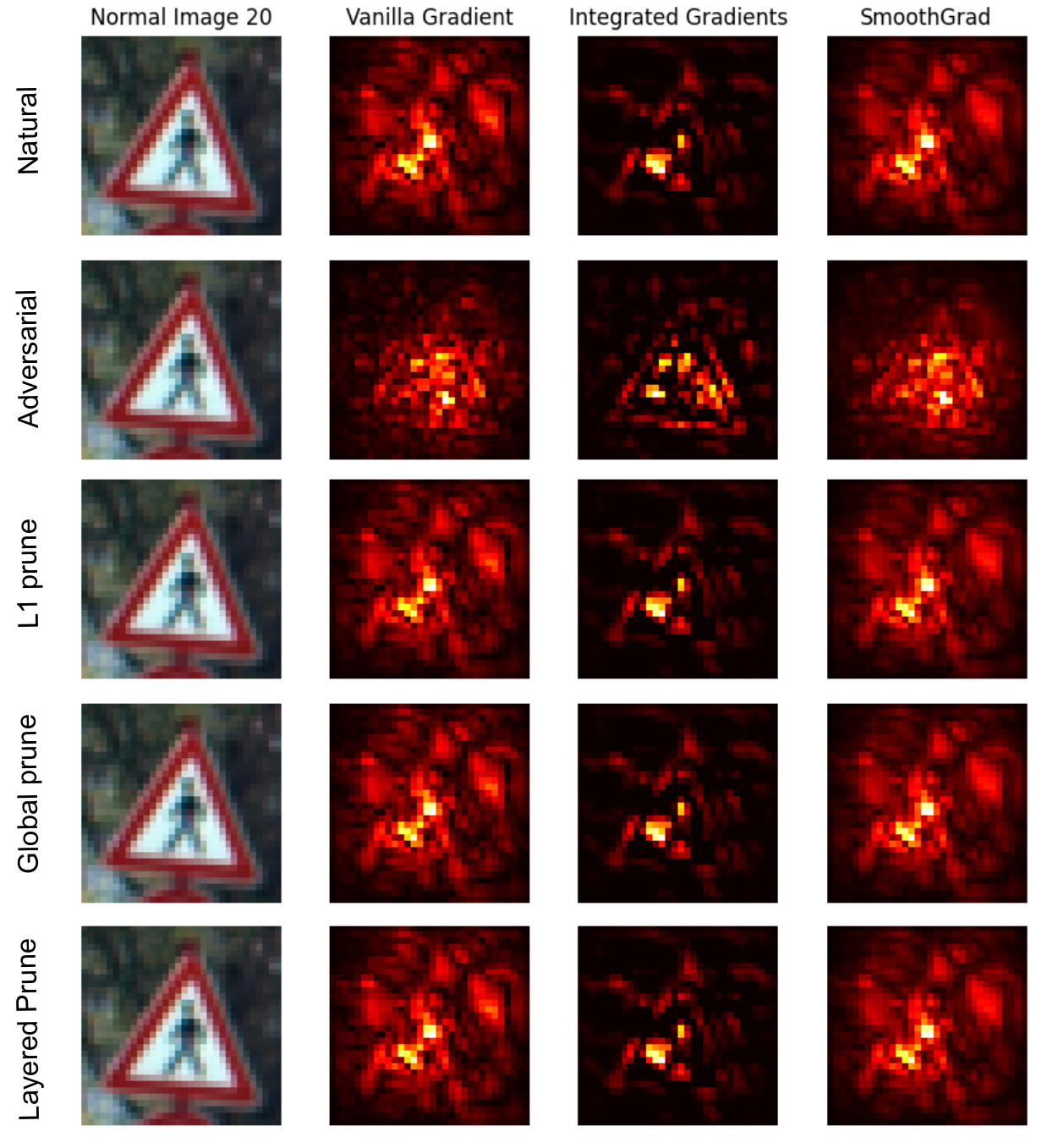}
    \caption{Saliency maps comparison for GTSRB dataset between natural training, adversarial training and post-train pruning without fine-tuning.}
    \label{fig:gtsrbadv1posttrainnoft}
\end{figure}

\begin{figure}[h]
    \centering
    \includegraphics[width=0.85\linewidth]{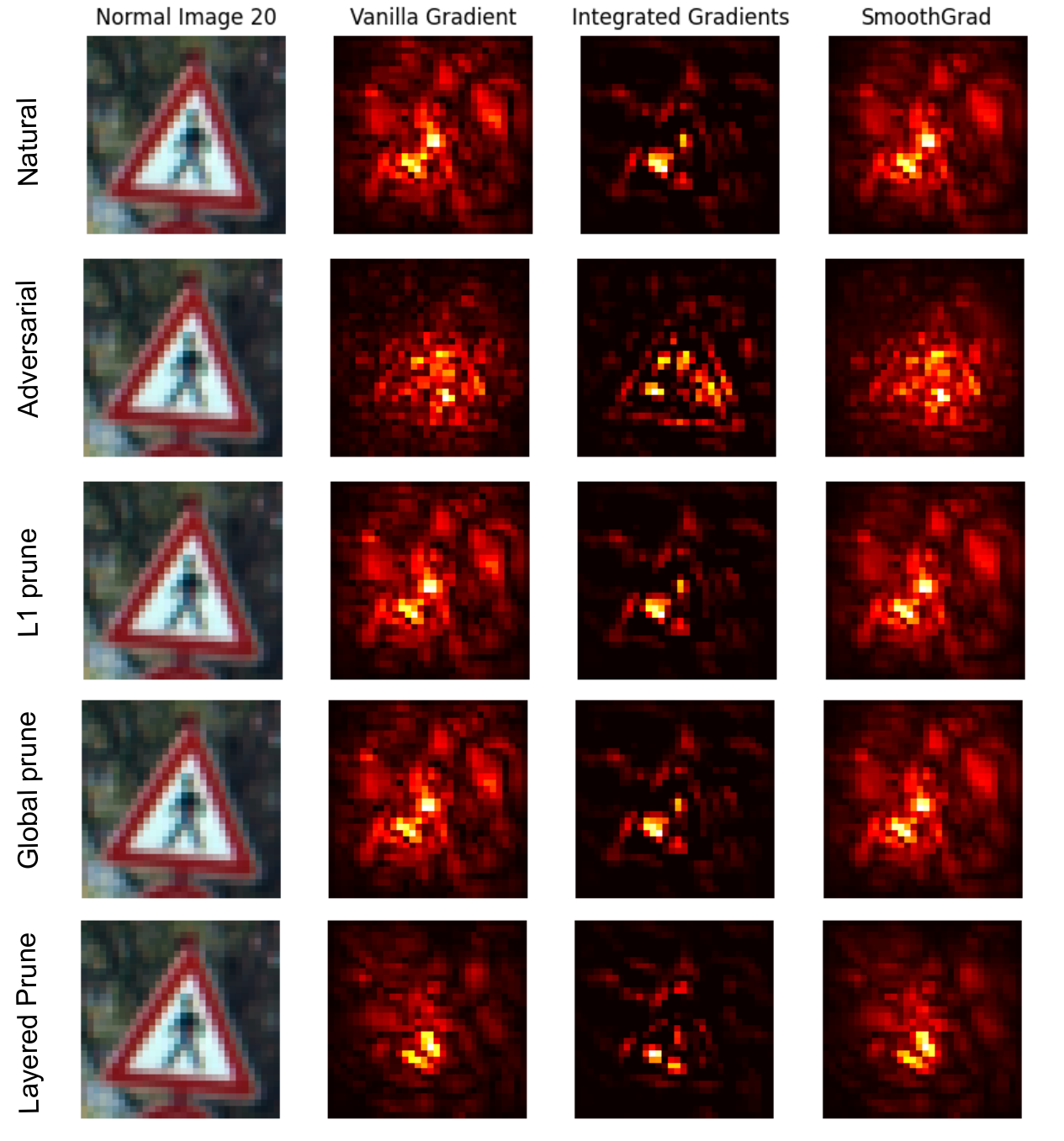}
    \caption{Saliency maps comparison for GTSRB dataset between natural training, adversarial training and post-train pruning with fine-tuning.}
    \label{fig:gtsrbadv1posttrainyesft}
\end{figure}

Figure \ref{fig:gtsrbadv1posttrainnoft} and Figure \ref{fig:gtsrbadv1posttrainyesft} shows the saliency maps comparison between naturally trained, adversarially trained and pruned models without and with fine-tuning respectively. 

\end{document}